\pdfoutput=1

\PassOptionsToPackage{xcdraw,table}{xcolor}
\PassOptionsToPackage{hyphens}{url}

\documentclass[11pt]{article}

\usepackage[final]{acl}

\usepackage{times}
\usepackage{latexsym}
\usepackage[T1]{fontenc}
\usepackage[utf8]{inputenc}
\usepackage{microtype}
\usepackage{inconsolata}
\usepackage{graphicx}
\usepackage{pifont}
\usepackage{booktabs}
\usepackage{xspace}
\usepackage{caption}
\usepackage{subcaption}
\usepackage{cleveref}
\usepackage{soul}
\usepackage[colorinlistoftodos]{todonotes}

\crefformat{section}{§#2#1#3}

\newcommand{\etox}{\textsc{etox}\xspace}
\newcommand{\detoxify}{\textsc{detoxify}\xspace}
\newcommand{\mutox}{\textsc{MuTOX}\xspace}
\newcommand{\mutoxasr}{\textsc{MuTOX-ASR}\xspace}

\newcommand{\cmark}{\ding{51}}
\newcommand{\xmark}{\ding{55}}

\definecolor{vega-orange}{HTML}{ff7f0e}
\definecolor{vega-blue}{HTML}{1f77bf}

\title{On the Role of Speech Data in Reducing Toxicity Detection Bias}

\author{
 \textbf{Samuel J.~Bell\textsuperscript{*\textdagger1}},
 \textbf{Mariano Coria Meglioli\textsuperscript{\textdagger1}},
 \textbf{Megan Richards\textsuperscript{\textdagger2}},
 \textbf{Eduardo Sánchez\textsuperscript{\textdagger1,3}},
\\
 \textbf{Christophe Ropers\textsuperscript{1}},
 \textbf{Skyler Wang\textsuperscript{4}},
 \textbf{Adina Williams\textsuperscript{1}},
\\
 \textbf{Levent Sagun\textsuperscript{\textdagger\textdaggerdbl1}},
 \textbf{Marta R.~Costa-jussà\textsuperscript{\textdagger\textdaggerdbl1}}
\\
 \textsuperscript{1}{Meta FAIR}
 \textsuperscript{2}{New York University}
\\
 \textsuperscript{3}{University College London}
 \textsuperscript{4}{McGill University}
\\ 
 \small{
    \textsuperscript{*}{First author}
    \textsuperscript{\textdagger}{Core contributor}
    \textsuperscript{\textdaggerdbl}{Senior author}
 }
\\
 \small{
   \textbf{Correspondence:} \href{mailto:sjbell@meta.com}{sjbell@meta.com}
 }
}

\usepackage{xcolor}

\begin{document}

\maketitle

\begin{abstract}
\textit{Text} toxicity detection systems exhibit significant biases, producing disproportionate rates of false positives on samples mentioning demographic groups.
But what about toxicity detection in \textit{speech}?
To investigate the extent to which text-based biases are mitigated by speech-based systems, we produce a set of high-quality group annotations for the multilingual \mutox dataset, and then leverage these annotations to systematically compare speech- and text-based toxicity classifiers.
Our findings indicate that access to speech data during inference supports reduced bias against group mentions, particularly for ambiguous and disagreement-inducing samples.
Our results also suggest that improving classifiers, rather than transcription pipelines, is more helpful for reducing group bias. 
We publicly release our annotations and provide recommendations for future toxicity dataset construction.
\end{abstract}

\noindent
\footnotesize
\textcolor{red}{\emph{Content warning: This paper contains toxic language that readers may find offensive or upsetting.}}
\normalsize

\section{Introduction}

With the growing prevalence of machine learning systems capable of processing and generating speech, there is rising interest in speech-aware toxicity detection \cite{costa-jussa-etal-2024-mutox,ghosh2022detoxy,nandwana2024voice,liu2024enhancing}.
Traditional cascaded approaches to speech toxicity detection use automated speech recognition (ASR) to convert speech to text, before applying a standard text classifier. 
This strategy has two main issues.
First, it eliminates rich prosodic and contextual information present in speech, which could degrade model performance. 
Second, text-based toxicity detection systems are well known to exhibit significant biases against minoritized groups \cite{dixon2018measuring,borkan2019nuanced}.
For instance, many systems are more likely to consider African American English (AAE) as toxic \cite{resende2024comprehensive}, while others denote the mere mention of identities such as ``gay'' and ``lesbian'' as toxic \cite{diasoliva2021fighting}.
Often, these issues are attributed to biases in the training data.
Because minoritized communities are overwhelmingly the subject of online toxicity \cite{dixon2018measuring,borkan2019nuanced}, classifiers misinterpret \emph{benign} group mentions as toxic, producing a disproportionate rate of false positives for marginalized groups \cite{dixon2018measuring}. 
Given these limitations, recent research has sought to develop toxicity classifiers that operate directly on speech. 

In this work, we perform a systematic comparison of speech-based and cascaded text-based toxicity detection systems.
Specifically, we hypothesize that access to speech audio provides useful contextual information, which could reduce false positives.
To investigate this, we produce a new set of annotations for a multilingual speech toxicity dataset, \mutox \cite{costa-jussa-etal-2024-mutox}, annotating for both toxicity and group mentions while also correcting automated transcripts.
To ensure consistent and accurate data, annotations were performed by the authors using a rigorous multi-stage process of cross-checking and discussion. 

We leverage these annotations to produce critical new insight into both the efficacy and biases of speech-based and text-based toxicity detection models.
\textbf{Our work reveals that incorporating speech data at inference time improves performance and reduces false positives on samples mentioning group identities, and eliminates false positives on ambiguous samples.}
Furthermore, we find that this bias is not the result of transcription error, but of the classifier itself.
We make our annotations publicly available to facilitate future research into the fairness and efficacy of speech-based toxicity detection.\footnote{\url{https://fb.me/mutox-group-annotations}}

\paragraph{Contributions.}
To summarize our main contributions, we: 
\begin{enumerate}
    \itemsep0em 
    \item Generate and release 1954 group annotations for speech toxicity detection fairness evaluations in English and Spanish; 
    \item Compare text- and speech-based toxicity detection systems, including detailed investigation of performance on ambiguous samples; 
    \item Isolate the role of transcription failure in text-based toxicity classifiers;
    \item Provide extensive analysis of the challenging ambiguity of toxicity annotation in speech.
\end{enumerate}

\begin{table*}[h!]
\caption{Annotation examples. English translations in \textcolor{gray}{gray}. See \cref{app:example-information} for corresponding \mutox IDs.}
\label{table:unambiguous-examples}
\footnotesize
\centering
\begin{tabular}{lp{8cm}ll}\hline
\toprule
& Transcript & Toxic & Group \\
\midrule
EN-1 & ``The Palestinian people does not exist'' & Yes (hate speech) & Racial or ethnic groups \\
EN-2 & ``I'm gonna have sex with this guy'' & No & Gender identities \\
ES-1 & ``Yo creo que la raza humana en general es una raza de mierda'' \newline \textcolor{gray}{``I believe that the human race in general is a shitty race''} & Yes (profanity) & - \\
ES-2 & ``Él era una persona muy mala, mató a muchos judíos'' \newline \textcolor{gray}{``He was a very bad person, killed many Jews''} &  No  & Religious groups \\
\bottomrule
\end{tabular}
\end{table*}

\begin{figure*}[ht!]
    \centering
     \begin{subfigure}[b]{0.20\textwidth}
        \centering 
        \includegraphics[trim={0 0 0 0},clip,height=3.75cm]{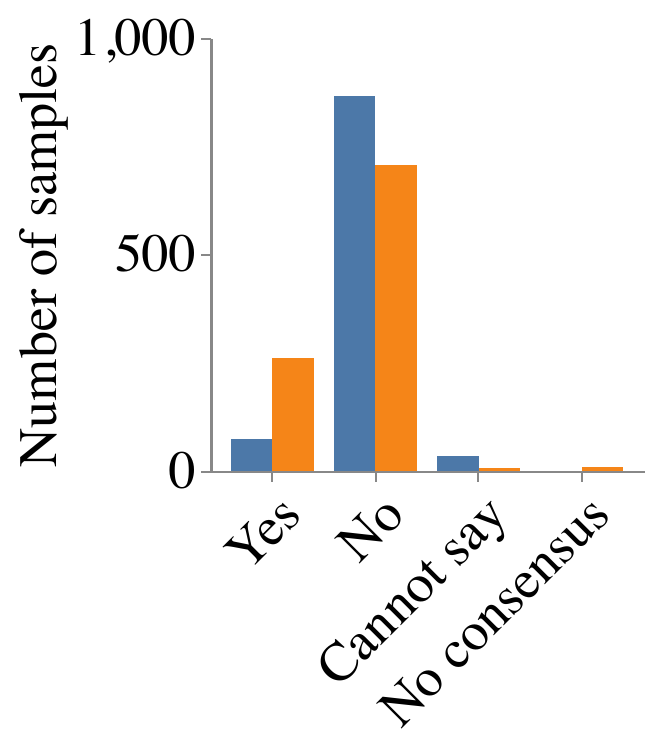}
        \vspace{-7pt}
        \caption{Toxicity}
        \label{fig:data-toxicity-by-lang}
    \end{subfigure}
    \hspace{1pt}
    \begin{subfigure}[b]{0.39\textwidth}
        \centering
        \hspace{6pt}
        \includegraphics[trim={0 0 0 0},clip,height=4cm]{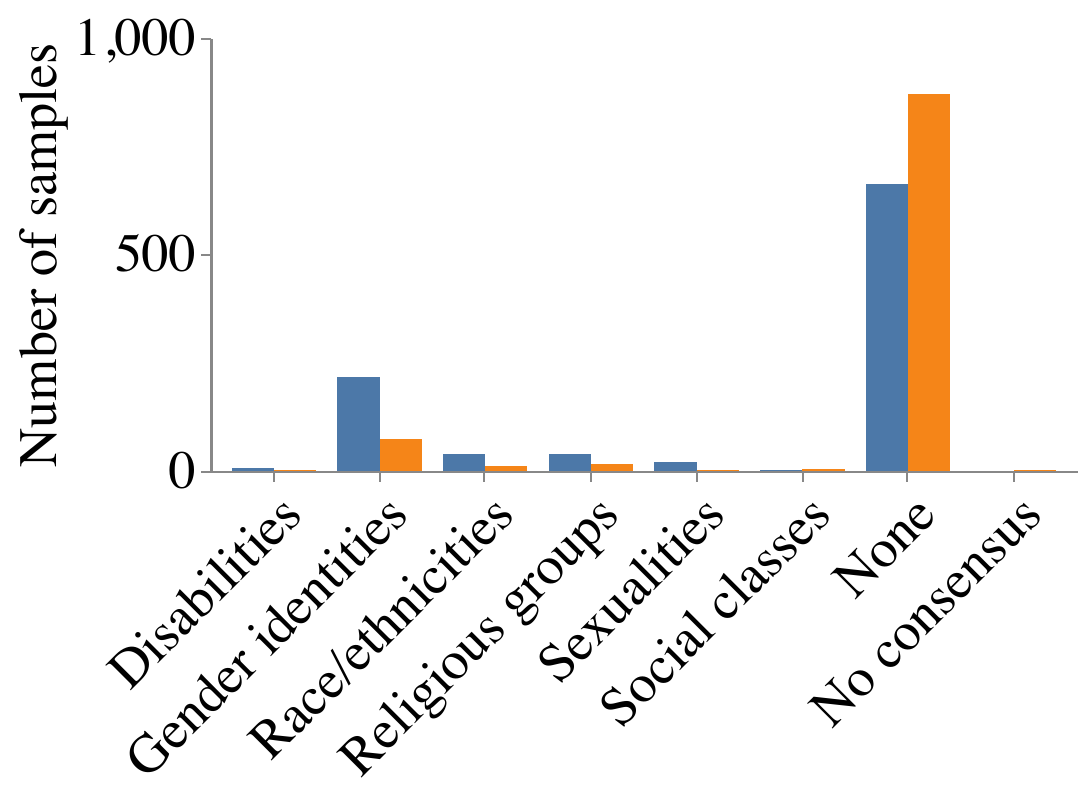}
        \caption{Groups}
        \label{fig:data-top-level-group-by-lang}
    \end{subfigure}
    \begin{subfigure}[b]{0.39\textwidth}
        \centering
        \includegraphics[trim={0 0 0 0},clip,height=4cm]{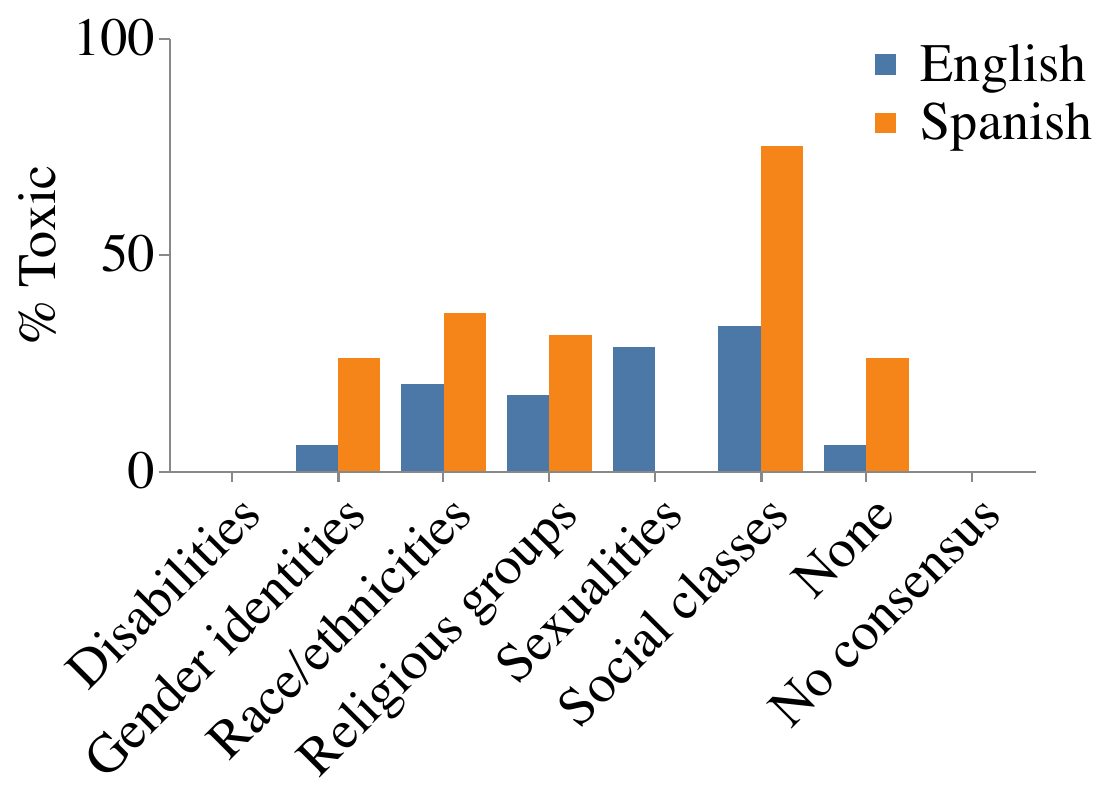}
        \caption{\% Toxicity}
        \label{fig:data-pct-toxicity-by-top-level-group-and-lang}
    \end{subfigure}
    \caption{(a) Number of samples marked as toxic (``Yes''), not toxic (``No''), impossible to decide (``Cannot say''), or where annotators could not reach consensus (``No consensus'') for \textcolor{vega-blue}{English} and \textcolor{vega-orange}{Spanish}. (b) Number of samples marked as mentioning or referring to a group. (c) Percentage of samples per group marked as toxic.}
    \label{fig:data-statistics}
\end{figure*}

\section{Background and related work}

\label{sec:background}

\subsection{Bias in toxicity detection}

Toxicity detection systems have long been known to exhibit significant biases (see \citealt{garg2023handling} for a review).
One major issue is the over-representation of certain identity markers in toxicity detection training data, often correlated with toxic content \cite{dixon2018measuring}.
For instance, models tend to conflate group mentions with toxicity, particularly for groups frequently targeted online, such as women, LGBTQ+ individuals, and minoritized racial, ethnic, or religious groups \cite{park2018reducing,borkan2019nuanced,diasoliva2021fighting}.
Models explicitly designed to detect anti-group bias also incorrectly associate group mentions with toxicity \citep{sahoo2022detecting}, unable to distinguish the \emph{use} of a term from a \emph{mention} \cite{gligoric2024nlp}.
Understanding how group mentions also bias speech toxicity classifiers is the key motivation of this work.

Toxicity classifiers have also been found to exhibit significant bias against AAE \citep{resende2024comprehensive}, partly due to annotator biases \cite{sap2022annotators,goyal2022your}.
Racial bias has similarly been observed in hate speech detection, which also suffers from the challenge of disambiguating genuinely hateful from reappropriated words  \cite{davidson2019racial,sap2019risk}.

Our work draws inspiration from Civil Comments \citep{borkan2019nuanced}, a text toxicity dataset with group annotations.
However, to better handle ambiguous cases, we opted to produce annotations ourselves rather than rely on crowd workers.

\subsection{Speech toxicity detection}

There is increasing interest in toxicity detection for speech data \cite{nandwana2024voice,liu2024enhancing}. 
The straightforward approach for constructing a speech-based toxicity detection system is a multi-stage pipeline, comprising an ASR stage followed by a text toxicity classification stage \cite{seamless2023seamlessm4t}.
Alternatively, models that operate directly on speech (e.g.~\citealt{costa-jussa-etal-2024-mutox}) typically utilize self-supervised speech encoders trained on large volumes of speech data, including wav2vec \cite{baevski2020wav2vec}, WavLM \cite{chen2022wavlm}, and SONAR \cite{duquenne2023sonar}.
Prior work in speech profanity detection suggests that models benefit from access to ``audio properties like pitch, emotions, [and] intensity'' \cite[p.~4]{gupta2022adima}.

While there are both monolingual \cite{ghosh2022detoxy} and multilingual \cite{gupta2022adima,costa-jussa-etal-2024-mutox} speech toxicity datasets, none are annotated with group information, precluding detailed analysis of bias against group mentions.

\subsection{Bias in speech systems}

Speech systems more broadly have been shown to exhibit biases in a range of contexts. 
For example, speech-based machine translation systems exhibit gender bias, such as by making gendered assumptions when translating between languages with and without grammatical gender \cite{costa-jussa-etal-2022-evaluating}.
The same phenomenon is present in speech-enabled large language models (LLMs), though its severity appears to be language-specific \cite{lin2024listen}.

Our work is closely connected to research exploring the biases of both ASR and self-supervised speech encoders such as SONAR (upon which \mutox is based).
Due to factors such as data imbalance \cite{garnerin2019gender}, ASR systems can exhibit gender bias \cite{tatman-2017-gender} and accent bias \cite{feng2021quantifying}, ultimately producing lower quality transcripts for certain groups of speakers.
SSL speech encoders also exhibit biases with respect to accent, age, and nationality \cite{lin2024listen}, though in contrast to ASR systems the composition of the pretraining data appears to have a limited effect \cite{boito2022study,meng2022dont}.
While speech \emph{data} may provide useful context that could reduce bias, speech \emph{pipelines} may add biases of their own, motivating our comparative study of text- and speech-based approaches.

\section{Annotating \mutox}

The foundational contribution of this work is a new, high-quality set of annotations for the \mutox test partition, allowing us to evaluate classifier bias against group mentions.
We believe this represents the first fairness audit dataset for multilingual speech toxicity detection.

\subsection{The \mutox dataset}
\mutox \cite{costa-jussa-etal-2024-mutox} is a large-scale, multilingual speech toxicity dataset covering 30 languages.
Each audio sample is accompanied by a text transcript produced by an open-source ASR model \cite{radford2023robust}.
For annotation tractability, we focus only on the English and Spanish test partitions, covering a total of 1954 samples. 

\subsection{Stage 1: Initial annotation}

We asked three annotators per language (all core contributors to this paper; see \cref{app:annotator-demographics}) to annotate the \mutox test set.
The annotators were all native-level proficient and spanned multiple language varieties (such as British and American English) to capture variety-specific interpretations.
Annotators used LabelStudio \cite{tkachenko2020labelstudio} with a custom interface (see \cref{app:annotation-interface}) to annotate for toxicity, group mentions, and automated transcript correctness. 

\paragraph{Toxicity.}
For toxicity, annotators were asked ``Does the audio contain toxicity?'' and presented with options for `Yes,'' ``No,'' or ``Cannot say,'' the latter indicating that the audio was unclear, truncated, or context-dependent.
Annotators were instructed to use the toxicity definition from the original \mutox annotation guidelines (see \cref{app:annotation-guidelines}), which defines toxicity as language which is ``typically considered offensive, threatening or harmful.''
This includes profanities and language related to physical violence, bullying, pornography, or hate speech.

\begin{table*}[t!]
\caption{Overview of the four toxicity detection systems. During training, all neural network models are trained on text, but only \mutoxasr and \mutox are trained jointly with speech data. At inference time, only \mutox has access to raw speech, while all other models rely on ASR text only.}
\label{table:model-details}
\footnotesize
\centering
\begin{tabular}{llcccc}
\toprule
Model & Type & \multicolumn{2}{c}{Train} & \multicolumn{2}{c}{Inference} \\
& & Text \ \ & Speech & ASR Text \ \ & Raw Speech \\
\midrule
\etox & Wordlist &  {\cellcolor{gray!25}} - & {\cellcolor{gray!25}} - & {\cellcolor{green!25}} \cmark & {\cellcolor{red!25}} \xmark \\ 
\detoxify & Neural network &  {\cellcolor{green!25}} \cmark & {\cellcolor{red!25}} \xmark & {\cellcolor{green!25}} \cmark & {\cellcolor{red!25}} \xmark \\ 
\mutoxasr & Neural network &  {\cellcolor{green!25}} \cmark & {\cellcolor{green!25}} \cmark & {\cellcolor{green!25}} \cmark & {\cellcolor{red!25}} \xmark \\ 
\mutox & Neural network &  {\cellcolor{green!25}} \cmark & {\cellcolor{green!25}} \cmark & {\cellcolor{green!25}} \cmark & {\cellcolor{green!25}} \cmark \\ 
\bottomrule
\end{tabular}
\end{table*}

\paragraph{Group mentions.}
For group annotation, annotators were asked ``Does the audio mention, or refer to (either explicitly or implicitly), any of the following?'' to which they could respond with one or more of ``Gender identities,'' ``Sexualities,'' ``Religious groups,'' ``Racial or ethnic groups,'' ``Disabilities,'' ``Social classes or socio-economic statuses,'' or ``None of the above.''
If any group was selected, annotators were asked then a follow-up about which specific group was mentioned.
For example, in the case of gender identities, they were asked ``Which gender identities are mentioned or referred to?'' with predefined options: ``Female, woman or girl,'' ``Male, man or boy,'' ``Nonbinary or gender non-conforming,'' and ``Transgender.''
Selectable groups were a superset of those used in Civil Comments \citep{borkan2019nuanced}, though annotators could provide free-text responses when the provided categories were insufficient.
See \cref{app:data-demographics} for the full list of groups annotated.

\paragraph{Transcript correction.}
After toxicity and group annotation, annotators were shown the audio's ASR transcript and asked ``Does this transcript match the audio?''
For the 21\% of samples where the transcript was inaccurate, annotators were required to correct it manually. \hfill \break

Before Stage 1, annotators conducted a pilot analysis of 20 samples (later discarded) to evaluate the interface and identify issues with the guidelines.
Annotators met frequently throughout Stage 1 to discuss problem cases and refine the guidelines, particularly regarding group annotation. 
In total, each annotator reviewed approximately 950 samples, spending approximately 30 to 45 seconds per sample.
See \cref{table:unambiguous-examples} for example annotations.

\subsection{Stage 2: Individual review}

Stage 1 responses were collated, and a majority vote was calculated for each sample.
For questions allowing multiple selections (e.g., group mentions), the majority vote was the set of options selected by at least two annotators.
Each annotator then independently reviewed the majority vote on a sample-by-sample basis.
Annotators flagged samples where they disagreed with the majority vote for further discussion in Stage 3, alongside all samples where there was complete disagreement.
Unflagged samples were assigned the majority vote as the final annotation.

\subsection{Stage 3: Group review}

Finally, annotators collectively reviewed all samples flagged during Stage 2, with the goal of sharing cultural knowledge and establishing consensus.
Discussions were conducted in language-specific groups, where the annotator who flagged a sample presented their rationale, followed by a group discussion.
Annotators were typically able to reach a consensus, but a ``No consensus'' label was occasionally assigned when annotators could not agree on a final label.
Note that while ``No consensus'' indicates that the annotators cannot agree on an outcome, ``Cannot say'' indicates that annotators \emph{agree} that toxicity could not be determined.
For example, all annotators might concur that the sample's interpretation depends on external context, such as the identity of the speaker or audience.
See \cref{fig:data-statistics} for a summary of the final annotations.

\section{The role of speech context}

\begin{figure*}[t!]
    \centering
     \begin{subfigure}[b]{0.325\textwidth}
        \centering
        \includegraphics[trim={0 0 0 0},clip,height=4cm]{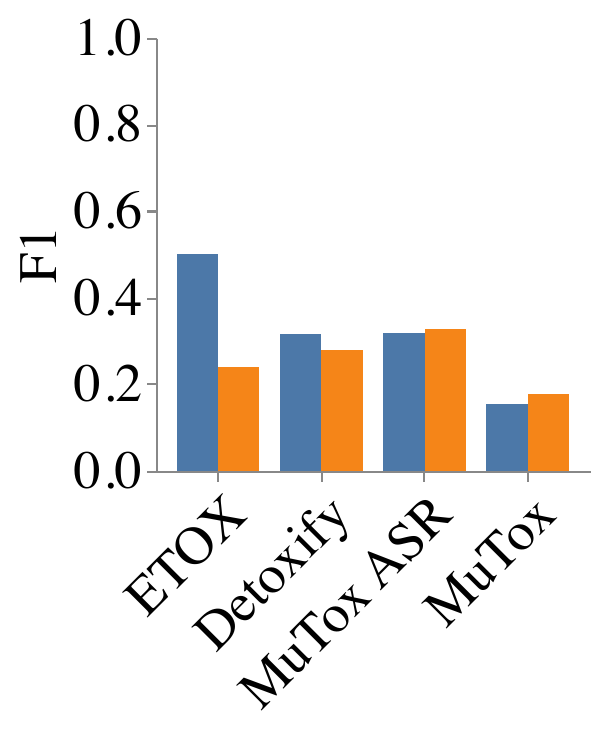}
        \caption{F-score}
        \label{fig:model-f1-by-top-level-group-all}
    \end{subfigure}
     \begin{subfigure}[b]{0.325\textwidth}
        \centering
        \includegraphics[trim={0 0 0 0},clip,height=4cm]{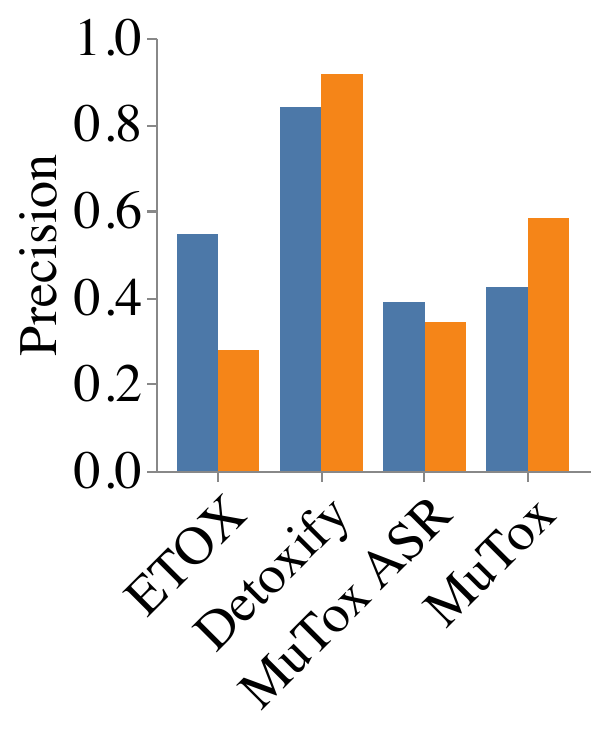}
        \caption{Precision}
        \label{fig:model-precision-by-top-level-group-all}
    \end{subfigure}
    \begin{subfigure}[b]{0.325\textwidth}
        \centering 
        \includegraphics[trim={0 0 0 0},clip,height=4cm]{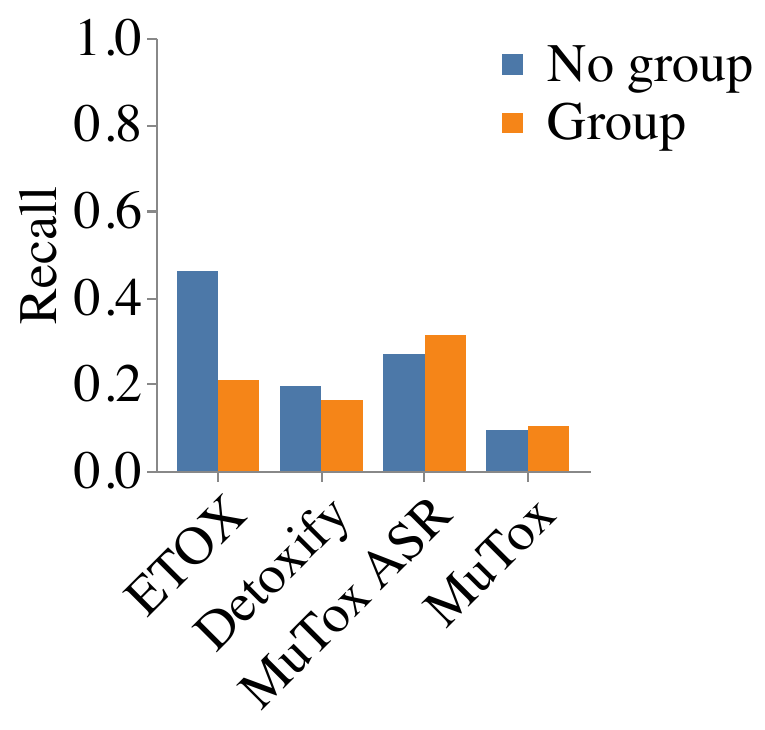}
        \caption{Recall}
        \label{fig:model-recall-by-top-level-group-all}
    \end{subfigure}
    \caption{(a) F-score, (b) precision, and (c) recall  of each classifier, for samples \textcolor{vega-orange}{with} and \textcolor{vega-blue}{without} group mentions. \etox and \detoxify show lower $F_1$-score when a group is mentioned, whereas \mutoxasr and \mutox show a slight increase. \mutox is the only classifier to increase both precision and recall when groups are mentioned.}
    \label{fig:model-f1-precision-recall-by-top-level-group}
\end{figure*}

We compare four representative toxicity classifiers to evaluate the utility of using speech data directly as opposed to cascaded ASR-based systems, and to isolate the role of speech during training from during inference (see \cref{table:model-details}).  

\subsection{Toxicity classifiers}

\textbf{\etox} \cite{costa-jussa-etal-2023-toxicity} is a text-only wordlist-based classifier that supports 200 languages.
While offering extensive coverage, it will only detect lexical toxicity and cannot account for context-dependent toxicity in polysemous words.

\textbf{\detoxify} ``multilingual'' \cite{hanu2020detoxify} is a text-only neural network that supports 7 languages and is trained on Wikipedia comments \cite{jigsaw-toxic-comment-classification-challenge} and Civil Comments \cite{borkan2019nuanced}, automatically translated using Google Translate. 

\textbf{\mutox} \cite{costa-jussa-etal-2024-mutox} is a multilingual neural network that supports 30 languages, trained on the \mutox dataset. 
\mutox is trained jointly on speech and text data encoded using SONAR \cite{duquenne2023sonar}.
At inference time, it operates on both speech audio and an accompanying text transcript.

\textbf{\mutoxasr} is similar to \mutox, but only has access to SONAR text embeddings at inference time. 
\mutoxasr can only access ASR transcripts but may benefit from improved representations developed during joint training.

\subsection{Methods}

For each of the four models, we extract predictions for every sample in the English and Spanish \mutox test sets.
For \etox, this is via lexical matching, whereas the model-based approaches all return a continuous toxicity score, subsequently binarized using a threshold. 
To ensure a fair comparison among all classifiers, the threshold was tuned on a per-language basis using the \mutox validation partition to match the precision of \etox.
We evaluate each model's performance using $F_1$-score, precision, and recall, and evaluate their bias against group mentions using false positive rate (FPR), following \citet{dixon2018measuring}.

\subsection{Results}

\begin{figure*}
    \centering
     \begin{subfigure}[b]{0.32\textwidth}
        \centering
        \includegraphics[trim={0 0 0 0},clip,height=4cm]{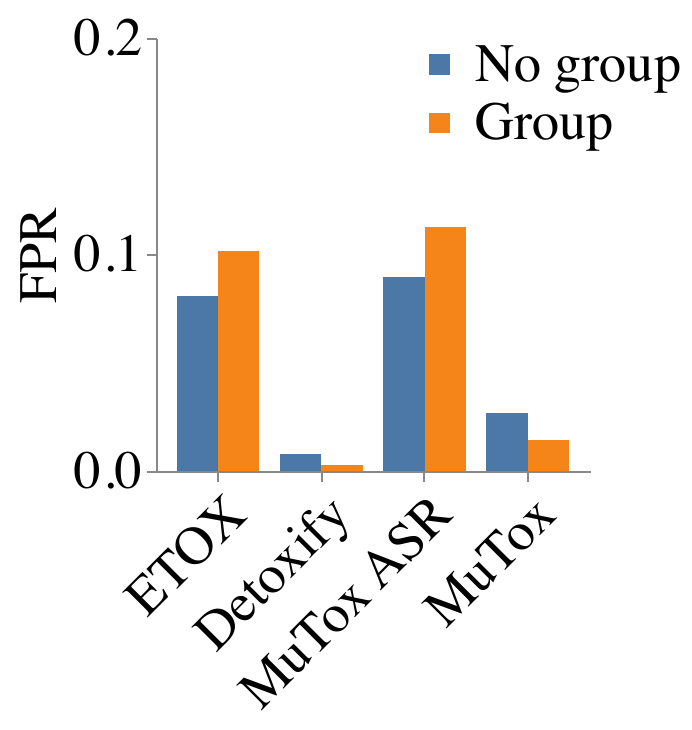}
        \caption{All samples}
        \label{fig:model-fpr-by-top-level-group-all}
    \end{subfigure}
    \hfill
    \begin{subfigure}[b]{0.32\textwidth}
        \centering
        \includegraphics[trim={0 0 0 0},clip,height=4cm]{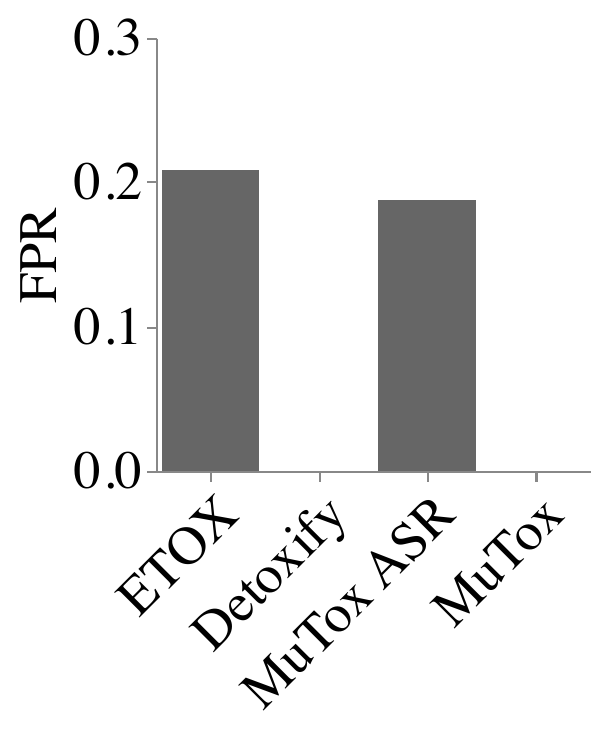}
        \caption{Ambiguous samples}
        \label{fig:model-fpr-nc-cs-only-no-group}
    \end{subfigure}
    \hfill
    \begin{subfigure}[b]{0.32\textwidth}
        \centering
        \includegraphics[trim={0 0 0 0},clip,height=4cm]{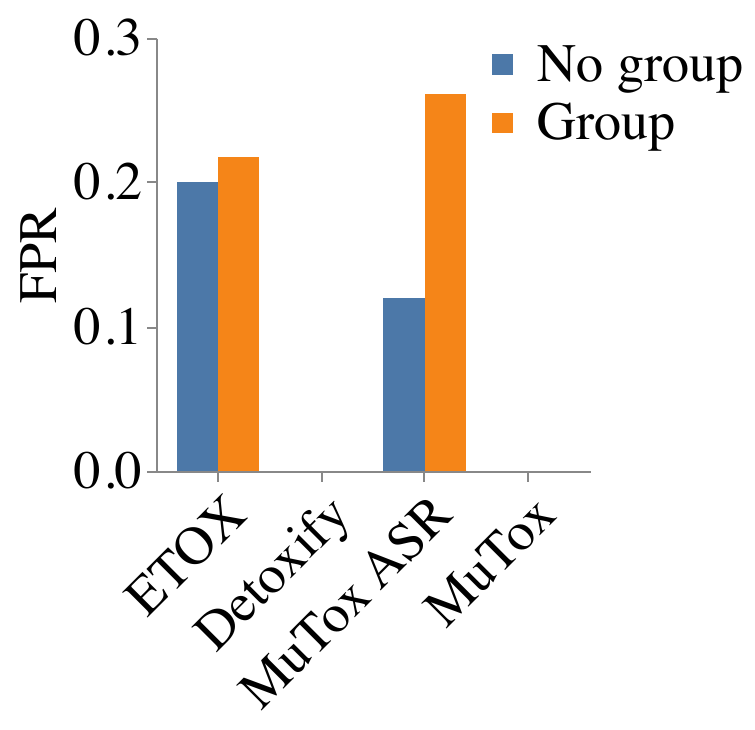}
        \caption{Ambiguous by group}
        \label{fig:model-fpr-nc-cs-only-by-top-level-group}
    \end{subfigure}
    \caption{(a) Classifier false positive rate (FPR) for samples \textcolor{vega-orange}{with}  and \textcolor{vega-blue}{without}  group mentions. (b) FPR of each classifier on samples annotators marked as ``Cannot say'' or ``No consensus.'' (c) FPR on ambiguous samples \textcolor{vega-orange}{with}  and \textcolor{vega-blue}{without}  group mentions. \detoxify and \mutox have an FPR of zero on ambiguous samples, while both \etox and \mutoxasr demonstrate increased FPR when ambiguous samples mention groups.}
    \label{fig:model-fpr}
\end{figure*}

Our evaluation reveals differences in the performance of speech-based and text-based toxicity detection models when sensitive groups are mentioned.
\Cref{fig:model-f1-precision-recall-by-top-level-group} shows that models relying solely on text (\etox, \detoxify) exhibit a reduced $F_1$-score. 
On the other hand, both models trained with speech data (\mutoxasr, \mutox) show a slight increase in $F_1$-score, but it is only the model with access to speech \emph{at inference time} (\mutox) that shows an increase across both precision and recall. 
Overall, while \mutox shows the worst $F_1$-score of all classifiers, its precision is markedly higher than \mutoxasr (given equivalent threshold tuning), which is particularly important in reducing false positives. 

Turning to FPR, \cref{fig:model-fpr-by-top-level-group-all} shows clear differences between classifiers. 
Wordlist-based \etox exhibits a high FPR that increases further when groups are mentioned, as does speech-trained \mutoxasr.
In contrast, \detoxify and \mutox both show low FPRs which decrease on group mentions.
While the high FPR for \etox is expected given the coarse nature of a wordlist, the differences between \mutoxasr (increase FPR on group mention) and \mutox (decrease on group mention) are particularly interesting. 
Both models are trained jointly with speech and text data, but only \mutox has access to speech data at inference time.
This suggests that if a model is trained on both speech and text, then making speech unavailable at inference time worsens anti-group bias. 
This may be due to an over-reliance on group mentions as cues in the absence of important speech context.

\begin{figure*}[ht!]
    \centering
     \begin{subfigure}[b]{0.32\textwidth}
        \centering
        \includegraphics[trim={0 0 0 0},clip,height=4cm]{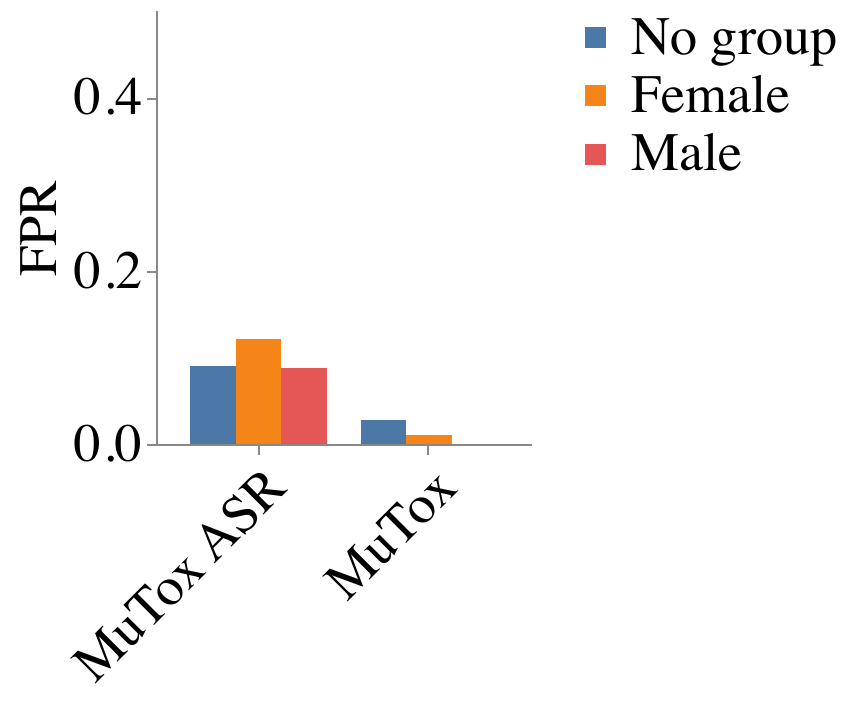}
        \caption{Gender identities}
        \label{fig:model-fpr-by-gender}
    \end{subfigure}
    \hfill
    \begin{subfigure}[b]{0.32\textwidth}
        \centering
        \includegraphics[trim={0 0 0 0},clip,height=4cm]{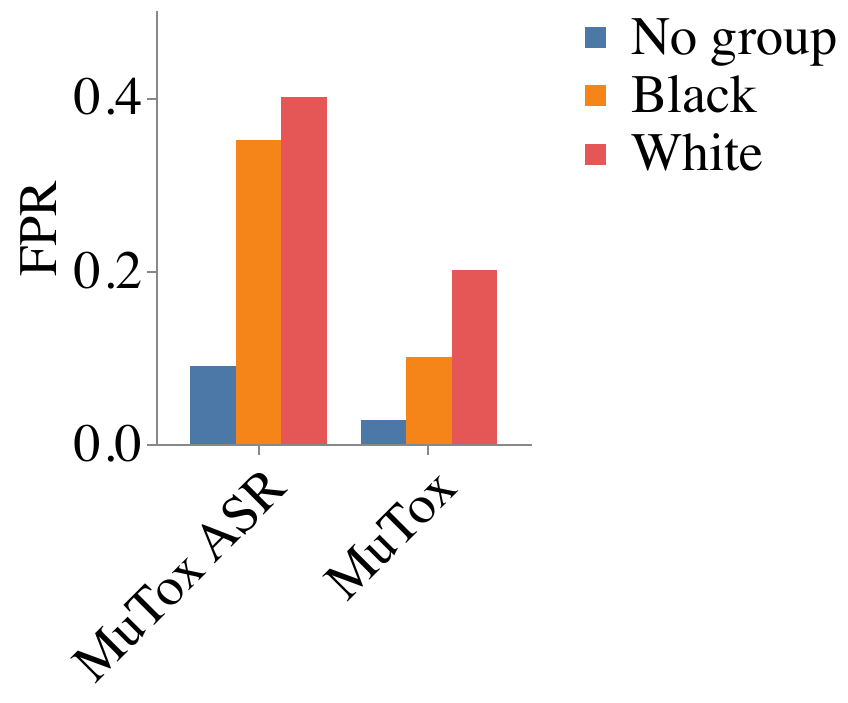}
        \caption{Racial or ethnic groups}
        \label{fig:model-fpr-by-race}
    \end{subfigure}
    \hfill
    \begin{subfigure}[b]{0.32\textwidth}
        \centering
        \includegraphics[trim={0 0 0 0},clip,height=4cm]{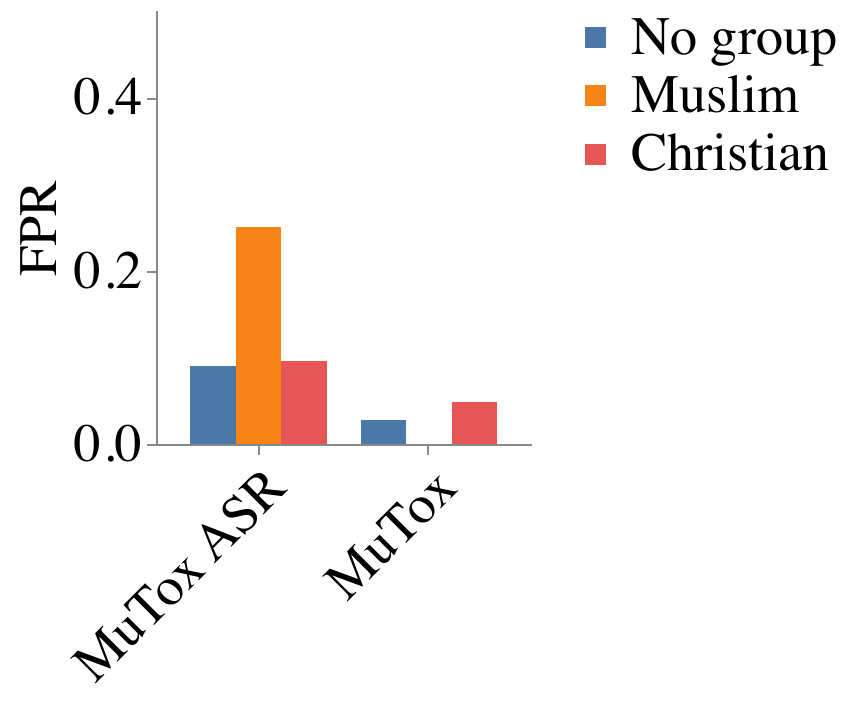}
        \caption{Religious groups}
        \label{fig:model-fpr-by-religion}
    \end{subfigure}
    \caption{False positive rate (FPR) of \mutox and \mutoxasr on samples mentioning specific (a) gender identities, (b) racial or ethnic groups, (c) religious groups. (a) Mutox ASR shows a higher FPR for samples mentioning women than for other samples, whereas \mutox's FPR decreases. (b) \mutoxasr shows a stronger bias against samples mentioning either White or Black people when compared to \mutox. (c) Similarly, \mutoxasr shows a stronger bias against religious group mentions than \mutox.}
    \label{fig:model-fpr-by-fine-grained-group}
\end{figure*}

Ambiguous samples---those labeled ``Cannot say'' or ``No consensus''---are a particular challenge for the wordlist-based \etox and \mutoxasr, while \detoxify and \mutox show an FPR of 0\% (\cref{fig:model-fpr-nc-cs-only-no-group}).
Once again, we see an increase in FPR when groups are mentioned for \mutoxasr (\cref{fig:model-fpr-nc-cs-only-by-top-level-group}).
This also supports our hypothesis that models trained to process speech but unable to leverage speech at inference time struggle to separate group mentions from toxicity. 

In \cref{fig:model-fpr-by-fine-grained-group}, we compare the FPR of \mutox and \mutoxasr on specific group mentions to further isolate the effect of incorporating speech data during inference.
With respect to gender (\cref{fig:model-fpr-by-gender}), \mutoxasr exhibits a higher FPR on samples mentioning women compared to samples with no group mentions, whereas \mutox shows a reduced FPR when samples mention either women or men.
Regarding race (\cref{fig:model-fpr-by-race}), both models show a higher FPR for samples mentioning Black people compared to no group mentions but unexpectedly show an even higher FPR for samples mentioning White people. 
As with gender, the increase in FPR for either group is reduced when incorporating speech during inference.
For samples mentioning religious groups (\cref{fig:model-fpr-by-religion}), \mutoxasr shows a higher FPR for samples mentioning Muslims compared to samples mentioning no group, while \mutox has an FPR of 0\% on these samples. 

Taken together, these results support our hypothesis that incorporating speech context during inference can help reduce toxicity detection failure and bias against certain groups, particularly for ambiguous or challenging samples. 
Notably, if a model is trained with speech data, our results suggest that it is important that the model \emph{operates} on speech at inference time to avoid leveraging neutral group mentions as shortcuts for toxicity.
That said, speech data is no panacea; speech-based models continue to exhibit biases in the form of increased FPR when certain groups are mentioned, suggesting systems should be deployed with caution. 

\section{Effect of transcription error}

One potential root cause of the failures observed in some cascaded ASR-based systems could be the ASR process.
In other words, to what extent are the performance differences between the text-based classifiers a result of transcription failures rather than biases in the classifier itself?
To address this question, we re-evaluate each classifier using the annotator-corrected transcripts. 

\begin{figure}
    \centering
     \begin{subfigure}[b]{0.48\columnwidth}
        \centering
        \includegraphics[trim={0 0 0 0},clip,height=4cm]{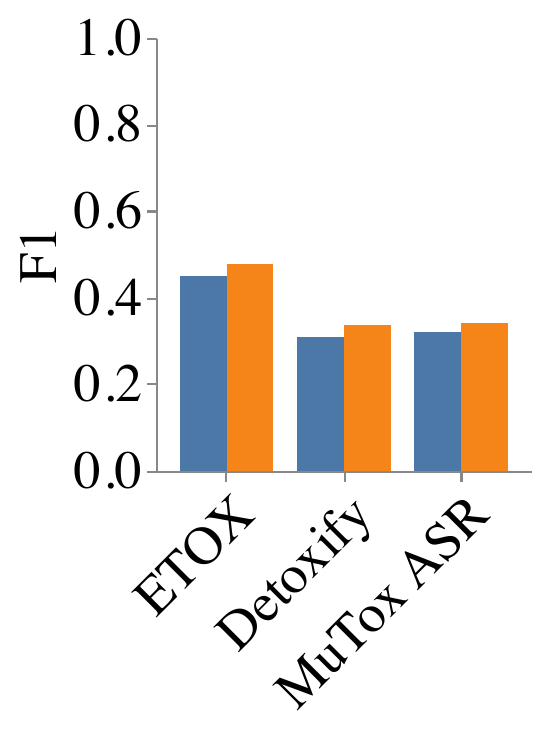}
        \caption{}
        \label{fig:model-f1-corrected-transcripts}
    \end{subfigure}
    \hfill
    \begin{subfigure}[b]{0.48\columnwidth}
        \centering
        \includegraphics[trim={0 0 0 0},clip,height=4cm]{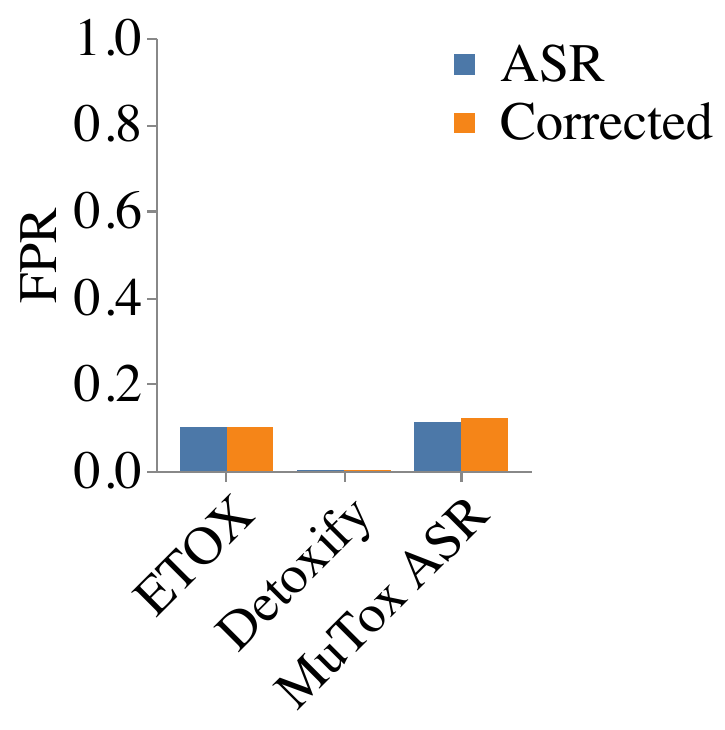}
        \caption{}
        \label{fig:model-fpr-group-mentions-corrected-transcripts}
    \end{subfigure}
    \caption{(a) $F_1$-score of cascaded ASR-based classifiers with \textcolor{vega-blue}{original ASR transcripts} and 
    \textcolor{vega-orange}{annotator-corrected transcripts}. (b) FPR on samples mentioning groups. Corrected transcripts only marginally improve model performance but have little to no impact on FPR.}
    \label{fig:model-performance-corrected-transcripts}
\end{figure}

In \cref{fig:model-f1-corrected-transcripts}, we observe that correcting the transcripts leads to a predictable improvement in the overall performance of the text-based classifiers.
At the same time, \cref{fig:model-fpr-group-mentions-corrected-transcripts} shows that the effect on the false positive rate (FPR) specifically for group mentions was minimal.
This suggests that transcription errors alone do not account for the observed biases in toxicity detection when group mentions are present and that refining transcription pipelines is unlikely to be a productive strategy for reducing bias in speech toxicity detection systems.

\begin{figure*}[ht!]
    \centering
     \begin{subfigure}[b]{0.32\textwidth}
        \centering
        \includegraphics[trim={0 0 0 0},clip,height=4cm]{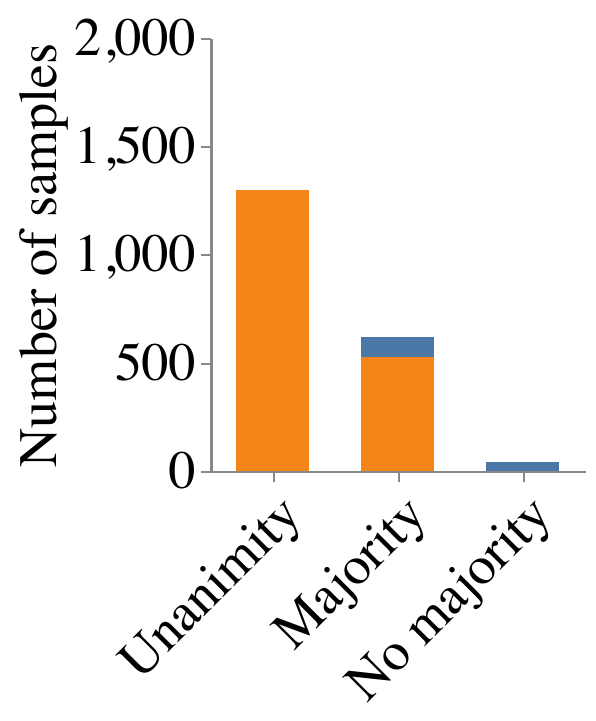}
        \caption{Toxicity}
        \label{fig:data-agreement-toxicity}
    \end{subfigure}
    \hfill
    \begin{subfigure}[b]{0.32\textwidth}
        \centering
        \includegraphics[trim={0 0 0 0},clip,height=4cm]{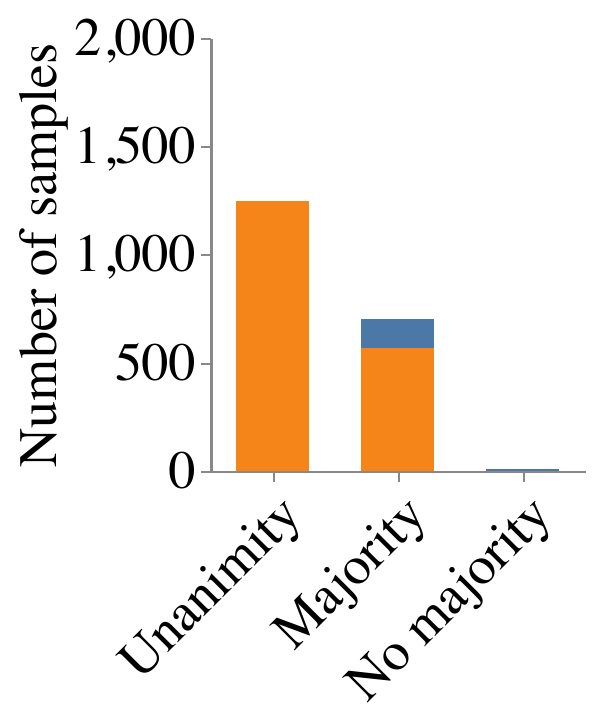}
        \caption{Group mention}
        \label{fig:data-agreement-topLevelGroup}
    \end{subfigure}
    \hfill
    \begin{subfigure}[b]{0.32\textwidth}
        \centering
        \includegraphics[trim={0 0 0 0},clip,height=4cm]{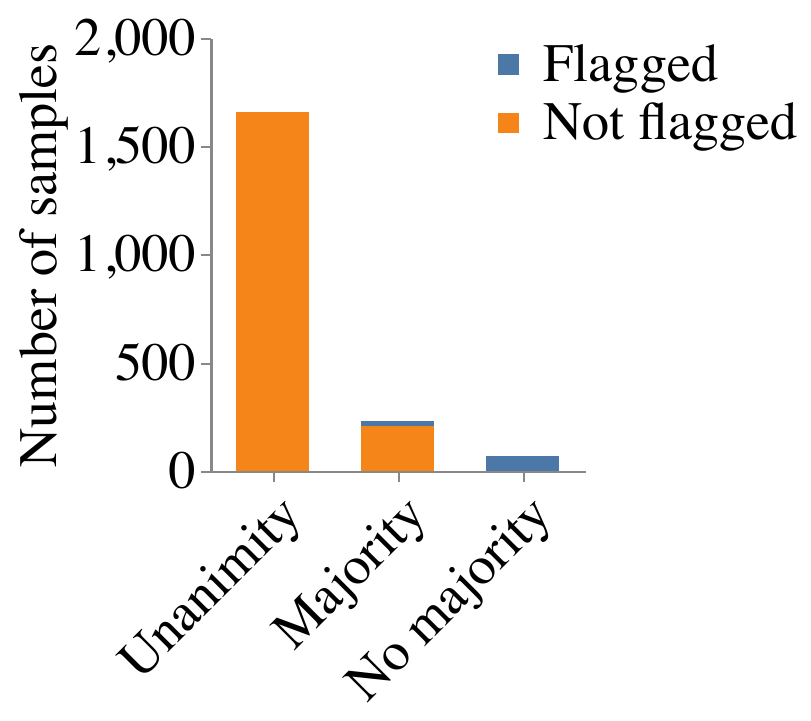}
        \caption{Transcription}
        \label{fig:data-agreement-transcription}
    \end{subfigure}
    \caption{Overview of inter-annotator agreement and review for (a) toxicity, (b) group annotation, and (c) transcription correction. Across all question types, annotators did not unanimously agree on a label for a sizeable proportion of the samples. Non-unanimous samples tended to have a majority vote, of which a reasonable fraction were flagged for Stage 2 review, alongside all samples lacking a majority.}
    \label{fig:annotator-flagging}
\end{figure*}

\section{Ambiguity in toxicity annotation}

\label{sec:ambiguity}

Our hypothesis that speech context can support less biased toxicity detection is predicated on the idea that toxicity itself is often highly subjective and context-dependent, making it hard to detect from outside of the initial conversation.
Indeed, our annotation process is a testament to this fact. 
While we intentionally designed the annotation process with multiple stages to support interactive discussion and consensus building, an analysis of annotator disagreement demonstrates the extent to which toxicity judgments can vary. 

After the first stage of annotation, annotators only unanimously agreed on toxicity in 66\% of samples.
For 32\% of samples, at least two annotators agreed, producing a majority vote, but for the 
remaining 2\% of samples, every annotator voted differently. 
A total of 7\% of samples were flagged for Stage 2 discussion (see \cref{fig:data-agreement-toxicity}).
These samples tended to be challenging to annotate, often requiring some degree of inference to determine what was left unsaid.
After review, annotators could not agree (``No consensus'') on 8 samples, whereas 40 samples resulted in a ``Cannot say'' (see \cref{fig:data-toxicity-by-lang}).

From the selection of flagged samples in \cref{table:ambiguous-examples}, we see that a variety of factors provoke discussion. 
For instance, annotators were unable to determine whether ``you fuckers'' (EN-3) was said in jest.
The toxicity of EN-4 depends on whether ``n***a'' is pejorative or a re-appropriated word; annotators were instructed not to draw inferences about speaker identity.
Sample ES-4 did not result in a consensus, as without further context, annotators were unable to determine the object referred to by ``monstruo'' (``monster'').
Annotators were conflicted about whether EN-5 refers to the speaker's viewpoint or to what others may say.
While annotators leaned towards marking this sample as toxic, disambiguating between genuine toxicity on the part of the speaker and quotations or reading passages (e.g., ES-3) was a persistent challenge, even in the case of a recognizable Bible passage (ES-5).

Annotators also exhibited similar levels of disagreement when annotating for group mentions (see \cref{fig:data-agreement-topLevelGroup}) despite our detailed and iterative shared guidelines.
A particular challenge for annotators was identifying whether certain group mentions corresponded to the category ``Racial or ethnic groups,'' as speakers rarely disambiguate between nationalities, ethnicities, or linguistic groups.
Ultimately, annotators reached a consensus after extensive discussion for all but one sample (see \cref{fig:data-top-level-group-by-lang}). 

During transcription correction, annotators unanimously agreed more frequently—about 85\% of samples, with only 5\% requiring Stage 2 review. 
Annotators failed to reach a consensus on the correct transcription for 7 samples, highlighting the difficulty inherent in cascaded approaches.

\begin{table*}[t!]
\caption{Selected ambiguous examples that provoked debate. Most samples lack sufficient context to fully understand the meaning of the segment, even with audio information. English translations in \textcolor{gray}{gray}. \mutox IDs in \cref{app:example-information}.}
\label{table:ambiguous-examples}
\footnotesize
\centering
\begin{tabular}{lp{11cm}ll}\hline
\toprule
& Transcript & Toxic \\
\midrule
EN-3 & ``We are gonna have a talk, you and me. You fuckers'' & Cannot Say \\
EN-4 & ``He is your below average n***a. So what you need to do'' & Cannot say \\
EN-5 & ``Yeah, you just gotta stop being gay and God will be okay with you'' & Yes \\ 
ES-3 & ``Pero si yo lo venzo y lo mato, entonces seréis nuestros siervos'' \newline \textcolor{gray}{``But if I defeat him and kill him, then you will be our servants''} & No consensus \\
ES-4 & ``\ldots creó este monstruo en el Medio Oriente'' \newline \textcolor{gray}{``\ldots created this monster in the Middle East''} & Yes \\
ES-5 & ``Entreguen ahora a esos malvados de Gibea, para que los matemos y eliminemos así la maldad de Israel'' \newline \textcolor{gray}{``Now hand over those wicked people from Gibea, so that we can kill them and thus eliminate the wickedness of Israel''} & Yes \\
\bottomrule
\end{tabular}
\end{table*}

\section{Discussion \& Conclusion}

Leveraging our new, high-quality set of group annotations for the \mutox test partition, we compared the performance and biases of text- and speech-based toxicity classifiers. 
Our analysis revealed that models that make use of speech data during \emph{both} training and inference exhibit reduced FPR bias against group mentions.
For ambiguous samples, we found that models trained on speech but without speech access at inference time exhibit an increased FPR, suggesting that the multimodal models rely on spurious correlations when lacking an informative modality.
Finally, we found that improving the quality of automated transcripts does little to reduce bias in English and Spanish, but this may change with lower-resourced languages where ASR systems exhibit poorer performance \cite{pratap2023scalingspeechtechnology1000}.

\subsection{The importance of multimodality}

Speech is not simply spoken text—the two linguistic forms diverge in grammar, morphology, and register. As a richer medium \cite{daft1987message}, speech encodes more information that helps one better ascertain communicative intent.
As such, even when the ``words'' converge, prosodic cues---e.g., inflection, tone, etc.---and contextual cues---e.g., speaker identity, social setting, etc.---in speech can contribute to differences in how meaning is construed in each of the two modalities \cite{kraut1992task}. 
By illustrating the improved performance of toxicity classifiers when speech data is introduced at inference time, we build on a growing body of work that demonstrates performance payoffs when engaging in multimodal and multitask learning.

\subsection{Toxicity beyond social media}

Much existing research on toxicity detection focuses on social media content moderation as the primary use case.
As a result, toxicity detection datasets (e.g.~\citealp{borkan2019nuanced}) are often drawn from social media.
This narrow focus may neglect increasingly relevant applications.
For instance, with the general public's growing interaction with LLMs, it may be desirable to detect toxicity in generated responses, which may be orthogonal to determining whether the content itself is safe.
Similarly, ensuring that machine translation systems do not introduce additional toxicity beyond what is present in the source is another emerging challenge \citep{sharou2022taxonomy}.

In contrast to earlier datasets, the \mutox dataset is primarily extracted from ``raw web corpora'' \citep[p.~2]{costa-jussa-etal-2024-mutox}, representing a broader range of toxicity data.
While this introduces certain biases (see \cref{sec:limitations}), it reflects a positive shift toward evaluating toxicity in more diverse contexts beyond social media.
As discussed in \cref{sec:ambiguity}, it is \emph{already} challenging for annotators to ascertain toxicity after the fact.
As toxicity datasets expand to include novel application domains, new combinations of modalities (e.g.~\citealp{kiela2020hateful}), and additional languages, robust annotation will become increasingly important.

\subsection{Practical recommendations}

To support the development of future speech toxicity datasets, we offer a few practical suggestions based on our experience annotating \mutox.

\paragraph{Speech first.}
We recommend that annotators be instructed to focus principally on audio when evaluating speech toxicity.
While audio may be unclear, ASR systems frequently make errors as they attempt to fill in gaps.
During Stage 3 group review, many initially ambiguous samples became clearer when the original audio was considered.

\paragraph{Iterate and refine.}
Annotators should be encouraged to reference, discuss, and update a working set of annotation guidelines, particularly when dealing with edge cases.
For example, while proper nouns were considered gender identity mentions, \mutox's skew towards liturgical content (see \cref{sec:limitations}) prompted extensive discussions about assigning gender to religious figures.
Shared guidelines and regular discussion can improve annotation consistency, but there is rarely a single, definitive answer.
When relying on crowd workers, where annotations are typically conducted in a single pass and disagreements resolved via majority vote, these nuances may be erroneously dismissed as noise.

\paragraph{Avoid automation.}
Recent work has explored using LLMs for annotation (e.g. \citealt{kumar2024decoding}) and benchmarking (e.g. \citealt{ustun-etal-2024-aya}).
In inherently subjective and context-dependent tasks like toxicity detection, the majority of samples exhibit at least some form of ambiguity, with many samples requiring extensive discussion, consideration of possible interpretations, and understanding of historical and political context. 
Conducting annotation without human annotators in the loop is unlikely to adequately capture such intricacies. 

\section{Limitations}  

\label{sec:limitations}

The \mutox dataset comprises audio clips ranging from 2 to 8 seconds, often leading to truncated fragments.
While annotators were instructed to make \emph{small and reasonable} inferences when the truncated obvious was sufficiently predictable, the short clip length likely contributed to an inflated number of "Cannot say" responses.
Truncated clips remove much-needed context \cite{pavlopoulos2020toxicity,xenos2021toxicity}, also amplifying the challenge of determining whether a speaker was expressing genuine toxicity or merely reading or quoting someone else.
Recent work has suggested that models struggle to distinguish between counterspeech and harmful content \cite{gligoric2024nlp}, but our findings indicate that this issue also arises during the annotation process itself.
Disambiguating between cases of "Cannot say" due to truncation versus genuine ambiguity would be more feasible with longer audio fragments, potentially improving annotation reliability.

Annotators were also discouraged from drawing inferences about speaker demographics, so annotators would typically assign a ``Cannot say'' to re-appropriated words (e.g., in AAE).
This approach may skew the distribution of toxicity labels for certain dialects.
Annotators also observed a noticeable skew in the topic distribution across both the English and Spanish data, with several annotators remarking that a significant number of samples were fragments of Bible passages or religious sermons.
Furthermore, the scope of our group annotations, intended for auditing rather than training, is limited by the time-intensive nature of annotation, with coverage constrained to English and Spanish. 
Future work should expand the sample size, domain diversity, and language coverage, particularly for under-resourced languages \citep{pratap2023scalingspeechtechnology1000}, to better understand the broader impact of speech-based toxicity detection systems.

Due to the time-intensive nature of our iterative annotation process, this work only considers two languages, English and Spanish, both of which are higher-resourced and well-studied languages.
Further research is required to understand the role of speech data in toxicity detection for lower-resourced languages.
In particular, if ASR pipelines exhibit higher error rates in lower-resourced languages, then improving them might be a more productive strategy than our results for English and Spanish suggest. 

Finally, we have not included statistical hypothesis testing when discussing our findings. 
Our principal contributions in this work are to produce a high-quality dataset of group annotations for multilingual speech toxicity detection, and subsequently to use those annotations to explore how classifier biases vary. 
As a result, we consider this work to be more exploratory than confirmatory \cite{bell2021perspectives}, and as such statistical hypothesis testing may not be appropriate . 



\bibliography{main}

\begin{thebibliography}{43}
\providecommand{\natexlab}[1]{#1}

\bibitem[{Adams et~al.(2017)Adams, Sorensen, Elliott, Dixon, McDonald, Thain, and Cukierski}]{jigsaw-toxic-comment-classification-challenge}
CJ~Adams, Jeffrey Sorensen, Julia Elliott, Lucas Dixon, Mark McDonald, Nithum Thain, and Will Cukierski. 2017.
\newblock Toxic comment classification challenge.
\newblock \url{https://kaggle.com/competitions/jigsaw-toxic-comment-classification-challenge}.

\bibitem[{Baevski et~al.(2020)Baevski, Zhou, Mohamed, and Auli}]{baevski2020wav2vec}
Alexei Baevski, Yuhao Zhou, Abdelrahman Mohamed, and Michael Auli. 2020.
\newblock \href {https://proceedings.neurips.cc/paper/2020/hash/92d1e1eb1cd6f9fba3227870bb6d7f07-Abstract.html} {Wav2vec 2.0: {A} framework for self-supervised learning of speech representations}.
\newblock In \emph{Advances in {{Neural Information Processing Systems}}}, volume~33, pages 12449--12460. Curran Associates, Inc.

\bibitem[{Barrault et~al.(2025)Barrault, Chung, Meglioli, Dale, Dong, Duquenne, Elsahar, Gong, Heffernan, Hoffman, Klaiber, Li, Licht, Maillard, Rakotoarison, Sadagopan, Wenzek, Ye, Akula, Chen, El~Hachem, Ellis, Gonzalez, Haaheim, Hansanti, Howes, Huang, Hwang, Inaguma, Jain, Kalbassi, Kallet, Kulikov, Lam, Li, Ma, Mavlyutov, Peloquin, Ramadan, Ramakrishnan, Sun, Tran, Tran, Tufanov, Vogeti, Wood, Yang, Yu, Andrews, Balioglu, {Costa-juss{\`a}}, {\c C}elebi, Elbayad, Gao, Guzm{\'a}n, Kao, Lee, Mourachko, Pino, Popuri, Ropers, Saleem, Schwenk, Tomasello, Wang, Wang, Wang, and {SEAMLESS Communication Team}}]{seamless2023seamlessm4t}
Lo{\"i}c Barrault, Yu-An Chung, Mariano~Coria Meglioli, David Dale, Ning Dong, Paul-Ambroise Duquenne, Hady Elsahar, Hongyu Gong, Kevin Heffernan, John Hoffman, Christopher Klaiber, Pengwei Li, Daniel Licht, Jean Maillard, Alice Rakotoarison, Kaushik~Ram Sadagopan, Guillaume Wenzek, Ethan Ye, Bapi Akula, Peng-Jen Chen, Naji El~Hachem, Brian Ellis, Gabriel~Mejia Gonzalez, Justin Haaheim, Prangthip Hansanti, Russ Howes, Bernie Huang, Min-Jae Hwang, Hirofumi Inaguma, Somya Jain, Elahe Kalbassi, Amanda Kallet, Ilia Kulikov, Janice Lam, Daniel Li, Xutai Ma, Ruslan Mavlyutov, Benjamin Peloquin, Mohamed Ramadan, Abinesh Ramakrishnan, Anna Sun, Kevin Tran, Tuan Tran, Igor Tufanov, Vish Vogeti, Carleigh Wood, Yilin Yang, Bokai Yu, Pierre Andrews, Can Balioglu, Marta~R. {Costa-juss{\`a}}, Onur {\c C}elebi, Maha Elbayad, Cynthia Gao, Francisco Guzm{\'a}n, Justine Kao, Ann Lee, Alexandre Mourachko, Juan Pino, Sravya Popuri, Christophe Ropers, Safiyyah Saleem, Holger Schwenk, Paden Tomasello, Changhan Wang, Jeff Wang,
  Skyler Wang, and {SEAMLESS Communication Team}. 2025.
\newblock \href {https://doi.org/10.1038/s41586-024-08359-z} {Joint speech and text machine translation for up to 100 languages}.
\newblock \emph{Nature}, 637(8046):587--593.

\bibitem[{Bell and Kampman(2021)}]{bell2021perspectives}
Samuel~J. Bell and Onno~P. Kampman. 2021.
\newblock \href {https://doi.org/10.48550/arXiv.2104.08878} {Perspectives on machine learning from psychology's reproducibility crisis}.
\newblock \emph{Preprint}, arXiv:2104.08878.

\bibitem[{Boito et~al.(2022)Boito, Besacier, Tomashenko, and Est{\`{e}}ve}]{boito2022study}
Marcely~Zanon Boito, Laurent Besacier, Natalia~A. Tomashenko, and Yannick Est{\`{e}}ve. 2022.
\newblock \href {https://doi.org/10.21437/INTERSPEECH.2022-353} {A study of gender impact in self-supervised models for speech-to-text systems}.
\newblock In \emph{23rd Annual Conference of the International Speech Communication Association, Interspeech 2022}, pages 1278--1282. {ISCA}.

\bibitem[{Borkan et~al.(2019)Borkan, Dixon, Sorensen, Thain, and Vasserman}]{borkan2019nuanced}
Daniel Borkan, Lucas Dixon, Jeffrey Sorensen, Nithum Thain, and Lucy Vasserman. 2019.
\newblock \href {https://doi.org/10.1145/3308560.3317593} {Nuanced metrics for measuring unintended bias with real data for text classification}.
\newblock In \emph{Companion {{Proceedings}} of {{The}} 2019 {{World Wide Web Conference}}}, {{WWW}} '19, pages 491--500. Association for Computing Machinery.

\bibitem[{Chen et~al.(2022)Chen, Wang, Chen, Wu, Liu, Chen, Li, Kanda, Yoshioka, Xiao, Wu, Zhou, Ren, Qian, Qian, Wu, Zeng, Yu, and Wei}]{chen2022wavlm}
Sanyuan Chen, Chengyi Wang, Zhengyang Chen, Yu~Wu, Shujie Liu, Zhuo Chen, Jinyu Li, Naoyuki Kanda, Takuya Yoshioka, Xiong Xiao, Jian Wu, Long Zhou, Shuo Ren, Yanmin Qian, Yao Qian, Jian Wu, Michael Zeng, Xiangzhan Yu, and Furu Wei. 2022.
\newblock \href {https://doi.org/10.1109/JSTSP.2022.3188113} {{{WavLM}}: {L}arge-scale self-supervised pre-training for full stack speech processing}.
\newblock \emph{IEEE Journal of Selected Topics in Signal Processing}, 16(6):1505--1518.

\bibitem[{Costa-juss{\`a} et~al.(2022)Costa-juss{\`a}, Basta, and G{\'a}llego}]{costa-jussa-etal-2022-evaluating}
Marta~R. Costa-juss{\`a}, Christine Basta, and Gerard~I. G{\'a}llego. 2022.
\newblock \href {https://aclanthology.org/2022.lrec-1.230/} {Evaluating gender bias in speech translation}.
\newblock In \emph{Proceedings of the Thirteenth Language Resources and Evaluation Conference}, pages 2141--2147. European Language Resources Association.

\bibitem[{Costa-juss{\`a} et~al.(2024)Costa-juss{\`a}, Meglioli, Andrews, Dale, Hansanti, Kalbassi, Mourachko, Ropers, and Wood}]{costa-jussa-etal-2024-mutox}
Marta~R. Costa-juss{\`a}, Mariano Meglioli, Pierre Andrews, David Dale, Prangthip Hansanti, Elahe Kalbassi, Alexandre Mourachko, Christophe Ropers, and Carleigh Wood. 2024.
\newblock \href {https://aclanthology.org/2024.findings-acl.340} {{M}u{T}ox: {U}niversal multilingual audio-based toxicity dataset and zero-shot detector}.
\newblock In \emph{Findings of the Association for Computational Linguistics ACL 2024}, pages 5725--5734. Association for Computational Linguistics.

\bibitem[{Costa-juss{\`a} et~al.(2023)Costa-juss{\`a}, Smith, Ropers, Licht, Maillard, Ferrando, and Escolano}]{costa-jussa-etal-2023-toxicity}
Marta~R. Costa-juss{\`a}, Eric Smith, Christophe Ropers, Daniel Licht, Jean Maillard, Javier Ferrando, and Carlos Escolano. 2023.
\newblock \href {https://doi.org/10.18653/v1/2023.findings-emnlp.642} {Toxicity in multilingual machine translation at scale}.
\newblock In \emph{Findings of the Association for Computational Linguistics: EMNLP 2023}, pages 9570--9586. Association for Computational Linguistics.

\bibitem[{Daft et~al.(1987)Daft, Lengel, and Trevino}]{daft1987message}
Richard~L Daft, Robert~H Lengel, and Linda~Klebe Trevino. 1987.
\newblock Message equivocality, media selection, and manager performance: Implications for information systems.
\newblock \emph{MIS Quarterly}, pages 355--366.

\bibitem[{Davidson et~al.(2019)Davidson, Bhattacharya, and Weber}]{davidson2019racial}
Thomas Davidson, Debasmita Bhattacharya, and Ingmar Weber. 2019.
\newblock \href {https://doi.org/10.18653/v1/W19-3504} {Racial bias in hate speech and abusive language detection datasets}.
\newblock In \emph{Proceedings of the {{Third Workshop}} on {{Abusive Language Online}}}, pages 25--35. Association for Computational Linguistics.

\bibitem[{Dias~Oliva et~al.(2021)Dias~Oliva, Antonialli, and Gomes}]{diasoliva2021fighting}
Thiago Dias~Oliva, Dennys~Marcelo Antonialli, and Alessandra Gomes. 2021.
\newblock \href {https://doi.org/10.1007/s12119-020-09790-w} {Fighting hate speech, silencing drag queens? {{A}}rtificial intelligence in content moderation and risks to {{LGBTQ} }voices online}.
\newblock \emph{Sexuality \& Culture}, 25(2):700--732.

\bibitem[{Dixon et~al.(2018)Dixon, Li, Sorensen, Thain, and Vasserman}]{dixon2018measuring}
Lucas Dixon, John Li, Jeffrey Sorensen, Nithum Thain, and Lucy Vasserman. 2018.
\newblock \href {https://doi.org/10.1145/3278721.3278729} {Measuring and mitigating unintended bias in text classification}.
\newblock In \emph{Proceedings of the 2018 {{AAAI}}/{{ACM Conference}} on {{AI}}, {{Ethics}}, and {{Society}}}, {{AIES}} '18, pages 67--73. Association for Computing Machinery.

\bibitem[{Duquenne et~al.(2023)Duquenne, Schwenk, and Sagot}]{duquenne2023sonar}
Paul-Ambroise Duquenne, Holger Schwenk, and Beno{\^i}t Sagot. 2023.
\newblock \href {https://doi.org/10.48550/arXiv.2308.11466} {{{SONAR}}: {S}entence-level multimodal and language-agnostic representations}.
\newblock \emph{Preprint}, arXiv:2308.11466.

\bibitem[{Feng et~al.(2021)Feng, Kudina, Halpern, and Scharenborg}]{feng2021quantifying}
Siyuan Feng, Olya Kudina, Bence~Mark Halpern, and Odette Scharenborg. 2021.
\newblock \href {https://doi.org/10.48550/arXiv.2103.15122} {Quantifying bias in automatic speech recognition}.
\newblock \emph{Preprint}, arXiv:2103.15122.

\bibitem[{Garg et~al.(2023)Garg, Masud, Suresh, and Chakraborty}]{garg2023handling}
Tanmay Garg, Sarah Masud, Tharun Suresh, and Tanmoy Chakraborty. 2023.
\newblock \href {https://doi.org/10.1145/3580494} {Handling bias in toxic speech detection: A survey}.
\newblock \emph{ACM Computing Surveys}, 55(13s).

\bibitem[{Garnerin et~al.(2019)Garnerin, Rossato, and Besacier}]{garnerin2019gender}
Mahault Garnerin, Solange Rossato, and Laurent Besacier. 2019.
\newblock \href {https://doi.org/10.1145/3347449.3357480} {Gender representation in {{French}} broadcast corpora and its impact on {{ASR}} performance}.
\newblock In \emph{Proceedings of the 1st {{International Workshop}} on {{AI}} for {{Smart TV Content Production}}, {{Access}} and {{Delivery}}}, {{AI4TV}} '19, pages 3--9. Association for Computing Machinery.

\bibitem[{Ghosh et~al.(2022)Ghosh, Lepcha, Singh, Shah, and Umesh}]{ghosh2022detoxy}
Sreyan Ghosh, Samden Lepcha, Sakshi Singh, Rajiv~Ratn Shah, and Srinivasan Umesh. 2022.
\newblock \href {https://doi.org/10.21437/INTERSPEECH.2022-10752} {Detoxy: {A} large-scale multimodal dataset for toxicity classification in spoken utterances}.
\newblock In \emph{23rd Annual Conference of the International Speech Communication Association, Interspeech 2022}, pages 5185--5189. {ISCA}.

\bibitem[{Gligoric et~al.(2024)Gligoric, Cheng, Zheng, Durmus, and Jurafsky}]{gligoric2024nlp}
Kristina Gligoric, Myra Cheng, Lucia Zheng, Esin Durmus, and Dan Jurafsky. 2024.
\newblock \href {https://doi.org/10.18653/v1/2024.naacl-long.331} {{{NLP}} systems that can't tell use from mention censor counterspeech, but teaching the distinction helps}.
\newblock In \emph{Proceedings of the 2024 {{Conference}} of the {{North American Chapter}} of the {{Association}} for {{Computational Linguistics}}: {{Human Language Technologies}} ({{Volume}} 1: {{Long Papers}})}, pages 5942--5959. Association for Computational Linguistics.

\bibitem[{Goyal et~al.(2022)Goyal, Kivlichan, Rosen, and Vasserman}]{goyal2022your}
Nitesh Goyal, Ian~D. Kivlichan, Rachel Rosen, and Lucy Vasserman. 2022.
\newblock \href {https://doi.org/10.1145/3555088} {Is your toxicity my toxicity? {E}xploring the impact of rater identity on toxicity annotation}.
\newblock In \emph{Proceedings of the ACM on Human-Computer Interaction}, volume 6 (CSCW2). Association for Computing Machinery.

\bibitem[{Gupta et~al.(2022)Gupta, Sharon, Sawhney, and Mukherjee}]{gupta2022adima}
Vikram Gupta, Rini Sharon, Ramit Sawhney, and Debdoot Mukherjee. 2022.
\newblock \href {https://doi.org/10.1109/ICASSP43922.2022.9746718} {{{ADIMA}}: {{A}}buse detection in multilingual audio}.
\newblock In \emph{{{{IEEE International Conference}} on {{Acoustics}}, {{Speech}} and {{Signal Processing}} ({{ICASSP}} 2022)}}, pages 6172--6176. IEEE.

\bibitem[{Hanu(2020)}]{hanu2020detoxify}
Laura Hanu. 2020.
\newblock Detoxify.
\newblock \url{https://github.com/unitaryai/detoxify}.

\bibitem[{Kiela et~al.(2020)Kiela, Firooz, Mohan, Goswami, Singh, Ringshia, and Testuggine}]{kiela2020hateful}
Douwe Kiela, Hamed Firooz, Aravind Mohan, Vedanuj Goswami, Amanpreet Singh, Pratik Ringshia, and Davide Testuggine. 2020.
\newblock \href {https://proceedings.neurips.cc/paper/2020/hash/1b84c4cee2b8b3d823b30e2d604b1878-Abstract.html} {The hateful memes challenge: {D}etecting hate speech in multimodal memes}.
\newblock In \emph{Advances in {{Neural Information Processing Systems}}}, volume~33, pages 2611--2624. Curran Associates, Inc.

\bibitem[{Kraut et~al.(1992)Kraut, Galegher, Fish, and Chalfonte}]{kraut1992task}
Robert Kraut, Jolene Galegher, Robert Fish, and Barbara Chalfonte. 1992.
\newblock Task requirements and media choice in collaborative writing.
\newblock \emph{Human--Computer Interaction}, 7(4):375--407.

\bibitem[{Kumar et~al.(2024)Kumar, Sahay, Mazumder, Okur, Manuvinakurike, Beckage, Su, Lee, and Nachman}]{kumar2024decoding}
Shachi~H. Kumar, Saurav Sahay, Sahisnu Mazumder, Eda Okur, Ramesh Manuvinakurike, Nicole Beckage, Hsuan Su, Hung-yi Lee, and Lama Nachman. 2024.
\newblock \href {https://doi.org/10.48550/arXiv.2408.03907} {Decoding biases: {A}utomated methods and {{LLM}} judges for gender bias detection in language models}.
\newblock \emph{Preprint}, arXiv:2408.03907.

\bibitem[{Lin et~al.(2024)Lin, Lin, Yang, Lu, Chen, Kuan, and Lee}]{lin2024listen}
Yi-Cheng Lin, Tzu-Quan Lin, Chih-Kai Yang, Ke-Han Lu, Wei-Chih Chen, Chun-Yi Kuan, and Hung-Yi Lee. 2024.
\newblock \href {https://doi.org/10.1109/SLT61566.2024.10832317} {Listen and speak fairly: A study on semantic gender bias in speech integrated large language models}.
\newblock In \emph{{{IEEE Spoken Language Technology Workshop}} ({{SLT}} 2024)}, pages 439--446. IEEE.

\bibitem[{Liu et~al.(2024)Liu, Nandwana, Pylkk{\"o}nen, Heikinheimo, and McGuire}]{liu2024enhancing}
Joseph Liu, Mahesh~Kumar Nandwana, Janne Pylkk{\"o}nen, Hannes Heikinheimo, and Morgan McGuire. 2024.
\newblock \href {https://arxiv.org/abs/2406.10325v1} {Enhancing multilingual voice toxicity detection with speech-text alignment}.
\newblock In \emph{25th Annual Conference of the International Speech Communication Association, Interspeech 2024}, pages 4298--4302. {ISCA}.

\bibitem[{Meng et~al.(2022)Meng, Chou, Liu, and Lee}]{meng2022dont}
Yen Meng, Yi{-}Hui Chou, Andy~T. Liu, and Hung{-}yi Lee. 2022.
\newblock \href {https://doi.org/10.1109/ICASSP43922.2022.9747897} {Don't speak too fast: The impact of data bias on self-supervised speech models}.
\newblock In \emph{{IEEE} International Conference on Acoustics, Speech and Signal Processing ({ICASSP} 2022)}, pages 3258--3262. {IEEE}.

\bibitem[{Nandwana et~al.(2024)Nandwana, He, Liu, {Xiao Yu}, Shang, Du~Bois, McGuire, and Bhat}]{nandwana2024voice}
Mahesh Nandwana, Yifan He, Joseph Liu, {Xiao Yu}, Charles Shang, Eloi Du~Bois, Morgan McGuire, and Kiran Bhat. 2024.
\newblock \href {https://doi.org/10.1109/ICASSP48485.2024.10448289} {Voice toxicity detection using multi-task learning}.
\newblock In \emph{2024 {{IEEE International Conference}} on {{Acoustics}}, {{Speech}} and {{Signal Processing}} ({{ICASSP}} 2024)}, pages 331--335. {IEEE}.

\bibitem[{Park et~al.(2018)Park, Shin, and Fung}]{park2018reducing}
Ji~Ho Park, Jamin Shin, and Pascale Fung. 2018.
\newblock \href {https://doi.org/10.18653/v1/D18-1302} {Reducing gender bias in abusive language detection}.
\newblock In \emph{Proceedings of the 2018 {{Conference}} on {{Empirical Methods}} in {{Natural Language Processing}}}, pages 2799--2804. Association for Computational Linguistics.

\bibitem[{Pavlopoulos et~al.(2020)Pavlopoulos, Sorensen, Dixon, Thain, and Androutsopoulos}]{pavlopoulos2020toxicity}
John Pavlopoulos, Jeffrey Sorensen, Lucas Dixon, Nithum Thain, and Ion Androutsopoulos. 2020.
\newblock \href {https://doi.org/10.18653/v1/2020.acl-main.396} {Toxicity detection: Does context really matter?}
\newblock In \emph{Proceedings of the 58th Annual Meeting of the Association for Computational Linguistics}, pages 4296--4305, Online. Association for Computational Linguistics.

\bibitem[{Pratap et~al.(2024)Pratap, Tjandra, Shi, Tomasello, Babu, Kundu, Elkahky, Ni, Vyas, Fazel-Zarandi, Baevski, Adi, Zhang, Hsu, Conneau, and Auli}]{pratap2023scalingspeechtechnology1000}
Vineel Pratap, Andros Tjandra, Bowen Shi, Paden Tomasello, Arun Babu, Sayani Kundu, Ali Elkahky, Zhaoheng Ni, Apoorv Vyas, Maryam Fazel-Zarandi, Alexei Baevski, Yossi Adi, Xiaohui Zhang, Wei-Ning Hsu, Alexis Conneau, and Michael Auli. 2024.
\newblock \href {http://jmlr.org/papers/v25/23-1318.html} {Scaling speech technology to 1,000+ languages}.
\newblock \emph{Journal of Machine Learning Research}, 25(97):1--52.

\bibitem[{Radford et~al.(2023)Radford, Kim, Xu, Brockman, Mcleavey, and Sutskever}]{radford2023robust}
Alec Radford, Jong~Wook Kim, Tao Xu, Greg Brockman, Christine Mcleavey, and Ilya Sutskever. 2023.
\newblock \href {https://proceedings.mlr.press/v202/radford23a.html} {Robust speech recognition via large-scale weak supervision}.
\newblock In \emph{Proceedings of the 40th {{International Conference}} on {{Machine Learning}}}, pages 28492--28518. PMLR.

\bibitem[{Resende et~al.(2024)Resende, Nery, Benevenuto, Zannettou, and Figueiredo}]{resende2024comprehensive}
Guilherme~H. Resende, Luiz~F. Nery, Fabrício Benevenuto, Savvas Zannettou, and Flavio Figueiredo. 2024.
\newblock \href {https://arxiv.org/abs/2401.12720} {A comprehensive view of the biases of toxicity and sentiment analysis methods towards utterances with african american english expressions}.
\newblock \emph{Preprint}, arXiv:2401.12720.

\bibitem[{Sahoo et~al.(2022)Sahoo, Gupta, and Bhattacharyya}]{sahoo2022detecting}
Nihar Sahoo, Himanshu Gupta, and Pushpak Bhattacharyya. 2022.
\newblock \href {https://doi.org/10.18653/v1/2022.conll-1.10} {Detecting unintended social bias in toxic language datasets}.
\newblock In \emph{Proceedings of the 26th Conference on Computational Natural Language Learning (CoNLL)}, pages 132--143. Association for Computational Linguistics.

\bibitem[{Sap et~al.(2019)Sap, Card, Gabriel, Choi, and Smith}]{sap2019risk}
Maarten Sap, Dallas Card, Saadia Gabriel, Yejin Choi, and Noah~A. Smith. 2019.
\newblock \href {https://doi.org/10.18653/v1/P19-1163} {The risk of racial bias in hate speech detection}.
\newblock In \emph{Proceedings of the 57th {{Annual Meeting}} of the {{Association}} for {{Computational Linguistics}}}, pages 1668--1678. Association for Computational Linguistics.

\bibitem[{Sap et~al.(2022)Sap, Swayamdipta, Vianna, Zhou, Choi, and Smith}]{sap2022annotators}
Maarten Sap, Swabha Swayamdipta, Laura Vianna, Xuhui Zhou, Yejin Choi, and Noah~A. Smith. 2022.
\newblock \href {https://doi.org/10.18653/v1/2022.naacl-main.431} {Annotators with attitudes: How annotator beliefs and identities bias toxic language detection}.
\newblock In \emph{Proceedings of the 2022 Conference of the North American Chapter of the Association for Computational Linguistics: Human Language Technologies}, pages 5884--5906. Association for Computational Linguistics.

\bibitem[{Sharou and Specia(2022)}]{sharou2022taxonomy}
Khetam~Al Sharou and Lucia Specia. 2022.
\newblock \href {https://aclanthology.org/2022.eamt-1.20} {A taxonomy and study of critical errors in machine translation}.
\newblock In \emph{Proceedings of the 23rd {{Annual Conference}} of the {{European Association}} for {{Machine Translation}}}, pages 171--180. European Association for Machine Translation.

\bibitem[{Tatman(2017)}]{tatman-2017-gender}
Rachael Tatman. 2017.
\newblock \href {https://doi.org/10.18653/v1/W17-1606} {Gender and dialect bias in {Y}ou{T}ube's automatic captions}.
\newblock In \emph{Proceedings of the First {ACL} Workshop on Ethics in Natural Language Processing}, pages 53--59. Association for Computational Linguistics.

\bibitem[{Tkachenko et~al.(2020)Tkachenko, Malyuk, Holmanyuk, and Liubimov}]{tkachenko2020labelstudio}
Maxim Tkachenko, Mikhail Malyuk, Andrey Holmanyuk, and Nikolai Liubimov. 2020.
\newblock \href {https://github.com/HumanSignal/label-studio} {{Label Studio}: {D}ata labeling software}.
\newblock \url{https://github.com/HumanSignal/label-studio}.

\bibitem[{{\"U}st{\"u}n et~al.(2024){\"U}st{\"u}n, Aryabumi, Yong, Ko, D{'}souza, Onilude, Bhandari, Singh, Ooi, Kayid, Vargus, Blunsom, Longpre, Muennighoff, Fadaee, Kreutzer, and Hooker}]{ustun-etal-2024-aya}
Ahmet {\"U}st{\"u}n, Viraat Aryabumi, Zheng Yong, Wei-Yin Ko, Daniel D{'}souza, Gbemileke Onilude, Neel Bhandari, Shivalika Singh, Hui-Lee Ooi, Amr Kayid, Freddie Vargus, Phil Blunsom, Shayne Longpre, Niklas Muennighoff, Marzieh Fadaee, Julia Kreutzer, and Sara Hooker. 2024.
\newblock \href {https://doi.org/10.18653/v1/2024.acl-long.845} {Aya model: An instruction finetuned open-access multilingual language model}.
\newblock In \emph{Proceedings of the 62nd Annual Meeting of the Association for Computational Linguistics (Volume 1: Long Papers)}, pages 15894--15939. Association for Computational Linguistics.

\bibitem[{Xenos et~al.(2021)Xenos, Pavlopoulos, Androutsopoulos, Dixon, Sorensen, and Laugier}]{xenos2021toxicity}
Alexandros Xenos, John Pavlopoulos, Ion Androutsopoulos, Lucas Dixon, Jeffrey Sorensen, and Leo Laugier. 2021.
\newblock \href {https://doi.org/10.48550/arXiv.2111.10223} {Toxicity detection can be sensitive to the conversational context}.
\newblock \emph{Preprint}, arXiv:2111.10223.

\end{thebibliography}

\clearpage 

\appendix

\section{Annotator demographics}

\begin{table*}[b]
\caption{Examples of samples exhibiting group mentions presented to annotators for reference.}
\label{table:annotation-guidelines}
\footnotesize
\centering
\begin{tabular}{lp{5cm}p{6cm}}\hline
\toprule
Group mention & Example & Reason \\
\midrule
Yes & ``All men suck'' & Mentions men, as a group \\
Yes & ``The woman drank her tea'' & Mentions a specific woman \\
Yes & ``She crossed the road'' & Refers to a woman via a gendered pronoun \\
Yes & ``Their pronouns are they/them'' & Refers to a person using gender neutral pronouns \\
Yes & ``The guy killed his houseplant'' & Mentions a specific man \\
Yes & ``As a queer person, I feel \ldots'' & Mentions a specific queer person \\
Yes & ``Jimmy’s best friend is gay'' & Mentions a specific queer person \\
Yes & ``Ella habla con su esposo'' & Mentions a woman via a gendered pronoun and references a specific man \\
Yes & ``Llámala, por favor'' & Mentions a woman via a gendered pronoun \\
No & ``They went to the party together'' & Uses a gender-neutral plural pronoun \\
No & ``Ustedes son los mejores'' & Uses a gender-neutral plural pronoun \\
No & ``Alex went shopping'' & Gender neutral name \\
\bottomrule
\end{tabular}
\end{table*}

Samples were annotated by three annotators with native- or near-native-level proficiency in the sample’s language.
Near-native proficiency is to be understood as CEFR level C2.
Annotators self-reported demographic information is described below.

Annotators had a mean age of 32.2 years with a standard deviation of 6.4 years. 
Three annotators described their gender as male, two as female, and one as nonbinary. 
Annotators described their ethnicity variously as White, White/Hispanic, Hispanic, or Middle Eastern, and were located in either France, the United Kingdom, or the United States.
Among English annotators, two spoke American English and one spoke British English.
For Spanish annotators, one spoke Cuban Spanish, one European Spanish and one Rioplatense Spanish. 

\label{app:annotator-demographics}

\section{Annotation interface}
\label{app:annotation-interface}

\begin{figure*}[b]
    \centering
    \includegraphics[trim={0 0 0 0},clip,width=0.7\textwidth]{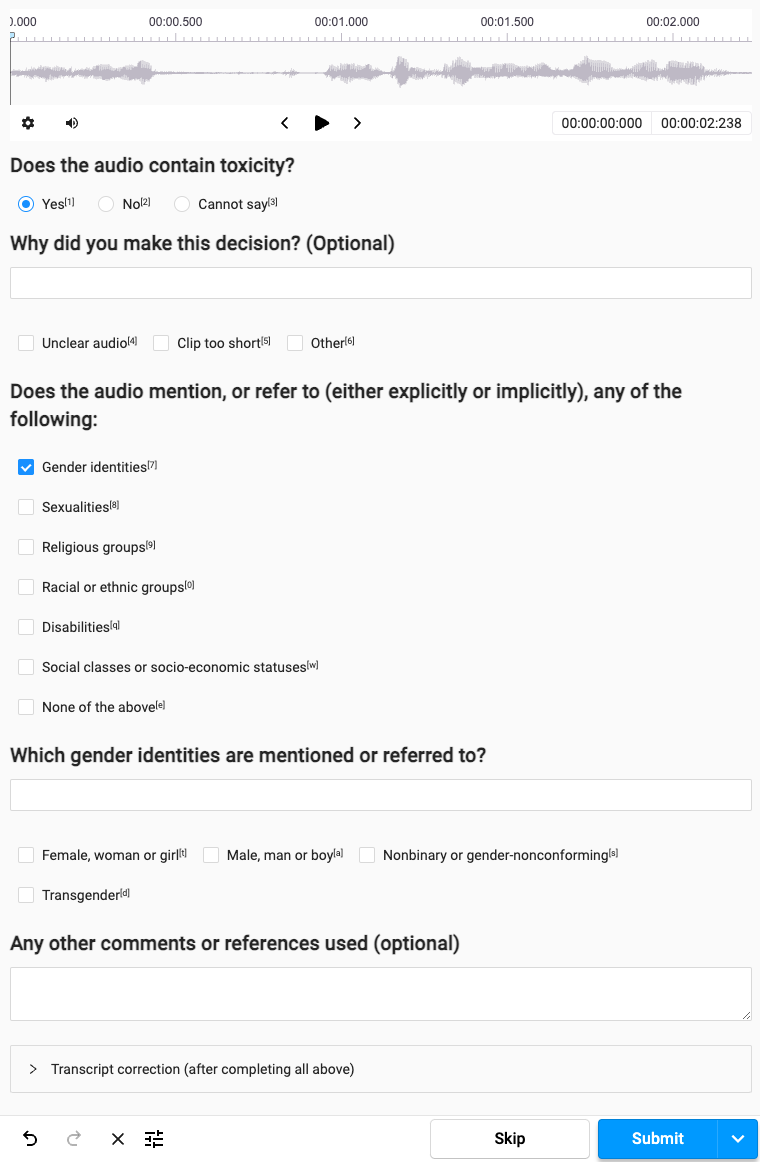}
    \caption{Annotation interface. Annotators could respond with free text if no checkbox was suitable.}
    \label{fig:annotation-interface}
\end{figure*}

\begin{figure*}[b]
    \centering
    \includegraphics[trim={0 0 0 0},clip,width=0.7\textwidth]{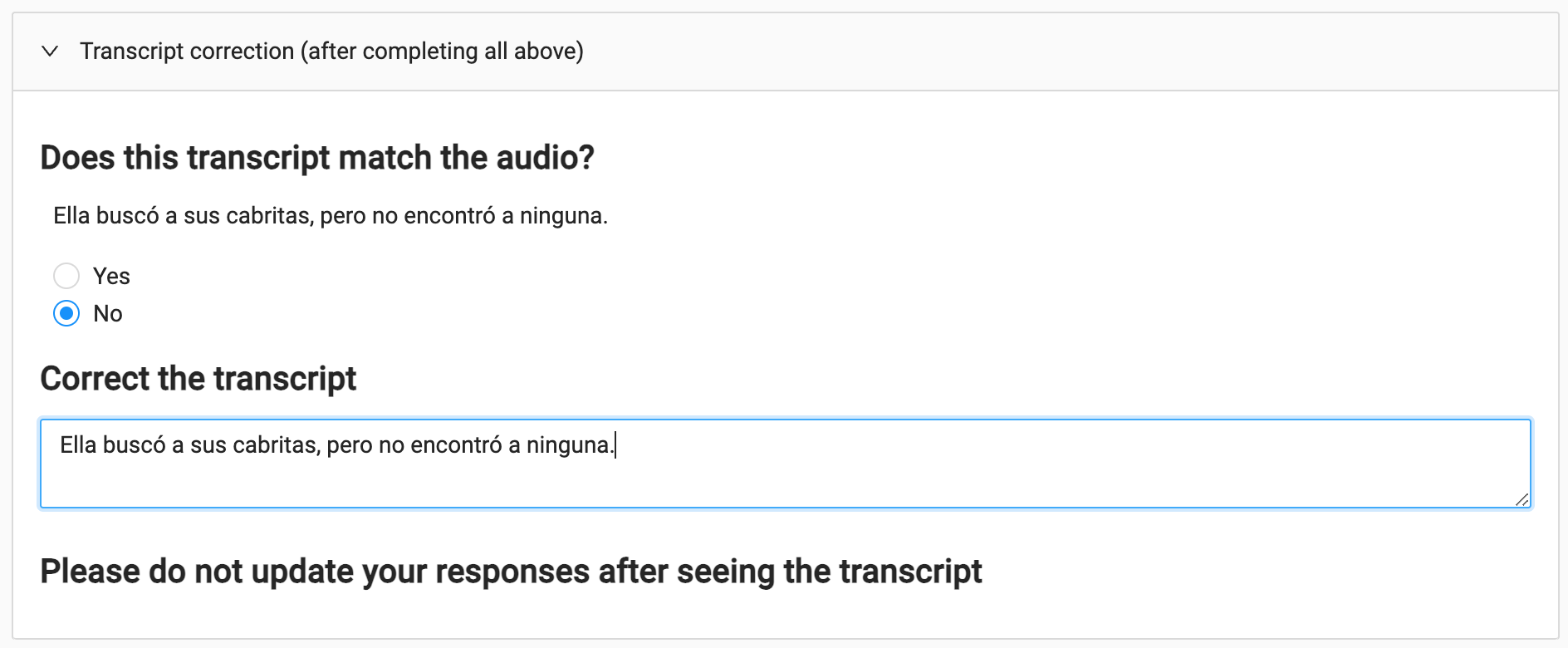}
    \caption{Transcription correction interface. Annotators were only asked to correct the transcript if they marked it as not matching the audio.}
    \label{fig:annotation-interface-transcription-correction}
\end{figure*}

See \cref{fig:annotation-interface} for the interface annotators used. 
Text transcripts were hidden until annotators had completed all other questions. 
See \cref{fig:annotation-interface-transcription-correction} for the expanded transcription correction interface.

\section{Annotation guidelines}
\label{app:annotation-guidelines}

\subsection{\mutox toxicity guidelines}

See \citet{costa-jussa-etal-2024-mutox} for full details, but we include relevant sections here.
\mutox defines toxicity as ``elements of language that are typically considered offensive, threatening, or harmful.''
\citeauthor{costa-jussa-etal-2024-mutox}'s definition spans:

\begin{itemize}
\item Profanities, defined as ``language that is regarded as obscene, repulsive, or excessively vulgar, as well as scatological.''
\item Hate speech, defined as ``language that is used to demean, disparage, belittle, or insult groups of people.''
\item Pornographic language, defined as ``language that refers to sexual acts or refers in a vulgar way to body parts typically associated with sexuality.''
\item Physical violence or bullying language, defined as ``language that is used to bully, threaten, silence individuals.''
\end{itemize}

During iterative discussion, annotators agreed that \emph{description} of violence, such as in a news report, should not be considered an example of the ``physical violence of bullying language'' category.

\subsection{Group guidelines}

We consider references to both groups as a whole and members of a group as mentions of that group.
This includes implicit references, such as using a proper noun, gendered pronoun, or grammatical gender markers (except where the gender is the default, such as using a masculine gender marker to refer to mixed groups of people in Spanish). 
Annotators collectively constructed the examples in \cref{table:annotation-guidelines} as a shared reference.

During iterative discussion, annotators also agreed on a number of small refinements to the annotation guidelines, which we include in the interest of transparency:
\begin{itemize}
\item References to \emph{all} religions, \emph{all} religious believers, etc., should not be considered mentions of a specific religious group. 
\item Annotating sexualities should be limited to mentions of sexual orientations, and exclude descriptions of sexual practices where a sexual orientation is not mentioned. 
\item While gender may sometimes be inferred from certain descriptions of sexualities, terms such as ``gay'' or ``lesbian'' should not be annotated as mentions of gender identities, unless gender is explicitly made apparent elsewhere in the sample.
\item The use of the term ``deformity'' should be interpreted as a group mention of physical disability, unless otherwise indicated from the context.
\end{itemize}

\begin{table*}[b]
\caption{Full \mutox IDs for examples mentioned in \cref{table:unambiguous-examples,table:ambiguous-examples}.}
\label{table:mutox-ids}
\footnotesize
\centering
\begin{tabular}{ll}\hline
\toprule
Text ID & \mutox ID \\
\midrule
EN-1 & 0253d9e34712d19727de811a\_2363424\_2366142 \\
EN-2 & e7d68d1bcb15dd5ca0baa6d6\_2394048\_2396766 \\
EN-3 & 25b2afe54ddab3f320478596\_1324992\_1329534 \\
EN-4 & e551701e4f0e2c64d58f4400\_1536000\_1540638 \\
EN-5 & 106325b4a23644d7b5aad341\_965280\_968286 \\
ES-1 & 255b54f0902d1919fbec7d86\_5690208\_5693598 \\
ES-2 & 248e4f212b0ffe4ad99bc7d8\_1683936\_1687326 \\
ES-3 & 5f575a5ab2945a3ffc6ab455\_266400\_271230 \\
ES-4 & e56d03f2445e80e9a864428b\_352896\_357150 \\
ES-5 & 89def0fab1f03646e53c5589\_376896\_382974 \\
\bottomrule
\end{tabular}
\end{table*}

\section{Group annotation results}
\label{app:data-demographics}

After Stage 3 review, the following groups were assigned to at least one or more samples for each identity category. 

\paragraph{Gender identities:} ``male, man or boy'', ``female, woman or girl'', ``transgender''

\paragraph{Sexualities:} ``homosexual, gay or lesbian'', ``queer'', ``bisexual'', ``heterosexual''

\paragraph{Religious groups:} ``christian'', ``jewish'', ``muslim''

\paragraph{Social classes or socio-economic statuses:} ``poverty'', ``working class'', ``agrarian'', ``upper class''

\paragraph{Racial or ethnic groups:} ``white'', ``african'', ``afghan'', ``russian'', ``jewish'', ``chinese'', ``black'', ``german'', ``palestinian'', ``english'', ``french'', ``indigenous american'', ``irish'', ``european'', ``indian'', ``ethiopian'', ``arab'', ``latino'', ``egyptian''

\section{Example information}
\label{app:example-information}

In the interest of brevity, samples mentioned in the main text are given a short identifier.
See \cref{table:mutox-ids} for the corresponding \mutox IDs for all samples in \cref{table:ambiguous-examples,table:unambiguous-examples}.

\end{document}